\documentclass{article}

\usepackage{iclr2024_conference,times}

\usepackage{amsmath,amsfonts,bm}

\def\eqref#1{equation~\ref{#1}}

\def\1{\bm{1}}

\DeclareMathAlphabet{\mathsfit}{\encodingdefault}{\sfdefault}{m}{sl}
\SetMathAlphabet{\mathsfit}{bold}{\encodingdefault}{\sfdefault}{bx}{n}

\usepackage{hyperref}
\usepackage{url}

\usepackage{wrapfig}
\usepackage{floatflt}
\usepackage{times}
\usepackage{epsfig}
\usepackage{graphicx}
\usepackage{amsmath}
\usepackage{amssymb}

\usepackage{subcaption}
\usepackage{graphicx}
\usepackage{amsmath}
\usepackage{amssymb}
\usepackage{booktabs}

\usepackage{lipsum}  

\usepackage{diagbox}
\usepackage{multirow}
\usepackage{comment}

\usepackage[inline]{enumitem}
\usepackage{arydshln}
\usepackage[dvipsnames]{xcolor}

\usepackage{algorithm} 
\usepackage{algorithmic}  
\usepackage[algo2e, linesnumbered,ruled,vlined]{algorithm2e}

\usepackage{float}

\SetCommentSty{mycommfont}

\SetKwInput{KwInput}{Input}                
\SetKwInput{KwOutput}{Output}              

\usepackage{paralist}
\usepackage{tikz-cd}
\usetikzlibrary{fit, shapes.geometric}

\usepackage{enumitem}

\newtheorem{theorem}{Theorem}
\newtheorem{example}[theorem]{Example}
\newtheorem{definition}[theorem]{Definition}
\newtheorem{remark}[theorem]{Remark}

\definecolor{simple}{HTML}{A50104}
\definecolor{comp}{HTML}{3C1518}
\definecolor{functor}{HTML}{3E8914}
\definecolor{source}{HTML}{240046}
\definecolor{target}{HTML}{5A189A}
\definecolor{free}{HTML}{9D4EDD}
\definecolor{appendix}{HTML}{990F3F}
\definecolor{train}{HTML}{D0F4DE}
\definecolor{function}{HTML}{e4c1f9}
\definecolor{input}{HTML}{fcf6bd}

\usepackage{caption}

\usepackage{paralist}

\usepackage{tikz}
\usetikzlibrary{calc}

\newcommand*{\tikzmk}[1]{\tikz[remember picture,overlay,] \node (#1) {};\ignorespaces}
\newcommand{\boxit}[1]{\tikz[remember picture,overlay]{\node[yshift=3pt,xshift=3pt,fill=#1,opacity=.25,fit={(A)($(B)+(.98\linewidth,.8\baselineskip)$)}] {};}\ignorespaces}

\usepackage[capitalize]{cleveref}
\crefname{section}{Sec.}{Secs.}
\Crefname{section}{Section}{Sections}
\Crefname{table}{Table}{Tables}
\crefname{table}{Tab.}{Tabs.}

\title{Pooling Image Datasets with Multiple Covariate Shift and Imbalance}

\author{Sotirios Panagiotis Chytas\thanks{Corresponding author. Email:  \texttt{chytas@wisc.edu} 
}\\ 
UW-Madison \\
\And
Vishnu Suresh Lokhande\\ 
UW-Madison \\ 
\And
Vikas Singh \\
UW-Madison\\
}

\iclrfinalcopy 
\begin{document}

\maketitle

\begin{abstract}
Small sample sizes are common in many disciplines, 
which necessitates pooling roughly similar datasets across 
multiple institutions to study weak but relevant 
associations between images and disease outcomes. Such 
data often manifest shift/imbalance in covariates 
(i.e., secondary non-imaging data). 
Controlling for such nuisance variables is 
common within standard statistical analysis, but 
the ideas do not directly apply to overparameterized models. 
Consequently, recent work has shown how strategies from 
invariant representation learning provides 
a meaningful starting point, but the current repertoire 
of methods is limited to accounting for shifts/imbalances in just a couple of covariates at a time. In this paper, we show how 
viewing this problem from the perspective of Category theory 
provides a simple and effective solution that completely avoids 
elaborate multi-stage training pipelines that would otherwise be 
needed. We show the effectiveness of this approach via 
extensive experiments on real datasets. Further, we 
discuss how this style of formulation offers a unified 
perspective on at least 5+ distinct 
problem settings, from self-supervised learning
to matching problems in 3D reconstruction. The code is available at \href{https://github.com/SPChytas/CatHarm}{https://github.com/SPChytas/CatHarm}.
\end{abstract}



\section{Introduction}\label{sec:intro}

\begin{wrapfigure}{r}{0.425\textwidth}
    \centering
    \vspace{-15pt}
    \includegraphics[width=0.42\textwidth]{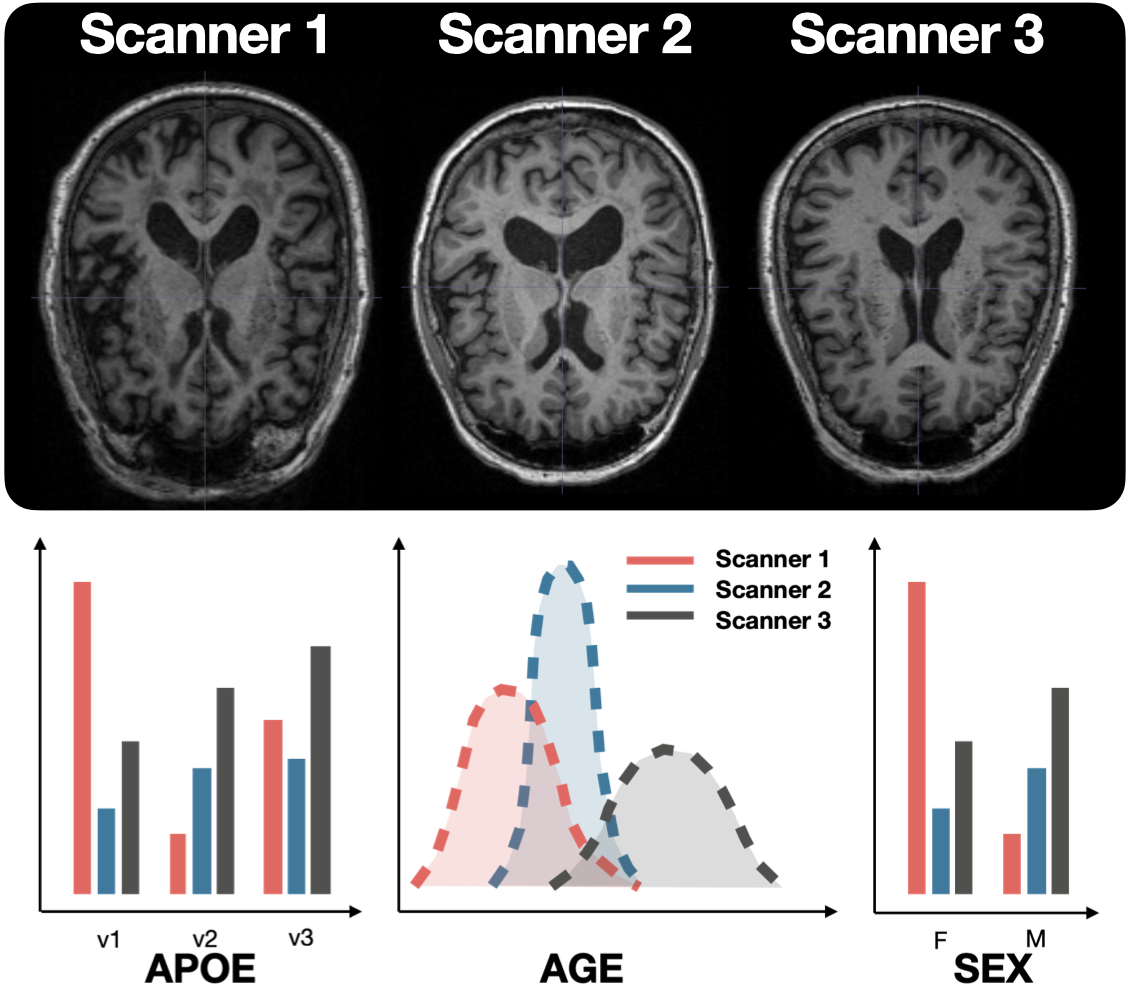}
    \vspace{-20pt}
    \caption{{\footnotesize{\bf Problem overview:} When pooling image datasets from different sources, there may be differences in the distribution of covariates. Covariates are secondary data for each individual that influences the images systematically. Top row shows MR images (say, different scanners). 
    Second row shows how the covariate distributions (genetic risk, age or gender) varies across scanners (but have a shared support). Learning representations from a pooled dataset in a manner such that covariate variations are accounted for, is challenging.}
    }
    \label{fig:overview}
    \vspace{-25pt}
\end{wrapfigure}

Sample sizes of medical imaging datasets at a single institution 
are often small due to many reasons. Acquiring thousands of images
can be infeasible due to budget/logistics. 
Also, if the scope of a study is narrow, only some individuals may be eligible due to inclusion criteria (e.g., have a specific genetic risk). To improve the statistical signal in  
retrospective analyses of existing 
 datasets, 
one option is to pool similar data 
across multiple sites \citep{enigma}. This offers a chance 
at discovering a real statistical 
effect, undetectable with small sample sizes.

Pooling data from multiple sites is common. A mature body of statistical literature 
(e.g., covariate matching \citep{stuart10,kim16}, meta-analysis \citep{enigma,rucker21})
describes best practices, and mechanisms to 
account for the effect of a nuisance variable on 
the response (or target label) are well known. 
This allows obtaining associations between relevant predictors and
the response/dependent variable of interest. 
For example, a standard analysis workflow may 
use {\em only} the images to predict a label (e.g., 
cognition) while controlling for nuisance 
variables such as race, gender or scanner type \citep{Penny2007StatisticalPM}.

\textbf{Controlling for covariates.} Deep neural network (DNN) models are now ubiquitous 
in medical image analysis \citep{unet, dl_medical}. 
But when using such models, systematic
mechanisms to deal with nuisance variables (or covariates), a key concern in data pooling, are still under development. 
Consider pooling MR images from two different sites where 
participant demographics distributions (e.g., age, sex etc) across sites are similar. If the scanners at the sites are different, we can include ``scanner'' or other continuous variables as a nuisance, and control for it within a general linear model. 
Features in the image $X$ deemed informative will explain variability in the response $Y$ {\em after} the 
variability due 
to ``age'' and ``site'' has been accounted for. In other words, 
the image data (and not the site-specific artifacts) should predict the response 
variable. 
Harmonization/pooling of imaging data in a way that accounts for (or removes)
the influence of non-imaging covariates is often limited to shallow pre-processing using additive/multiplicative batch corrections, e.g., in Combat \citep{johnson2007adjusting, combat_brain}.

{\bf Data pooling/harmonization and invariance.} In the context of representation learning, several results 
have identified the link between data pooling/harmonization and invariant representation learning or domain adaptation \citep{moyer2018invariant,moyer20}, which have 
been utilized for downstream tasks \citep{lokhande2020fairalm}. 
Initial approaches focused on image normalization techniques (e.g., histogram matching \cite{histogram}), which allowed controlling some variations in intensities across scanners or protocols. However, such approaches cannot systematically handle covariates.
Ideas based on invariant representations \citep{BloemReddy2020ProbabilisticSA,li2014learning,arjovsky-bottou-gulrajani-lopezpaz-2019} have  
allowed controlling for up to two covariates during 
representation learning. 
For example, one may ask that the model avoid using image features associated with two specific covariates (or in fairness terminology, sensitive attributes): age and site.
Nonetheless, models that can easily remove (or control for) the influence of multiple covariates remain limited and so, 
either Combat-based \citep{johnson2007adjusting} pre-processing is used 
\citep{combat_brain} or CycleGAN-based schemes \citep{cyclegan1,cyclegan2}
can harmonize the image data relative to a specific 
categorical variable (say, scanner). This problem is also relevant in the longitudinal setting \cite{Longitudinal_vae}.

{\bf The problem.} 
Different scanners can introduce systematic differences in the scans of the same person 
\citep{inter_scanner} and mitigation strategies continue
to be a topic of recent work \citep{relief}.
When there is
some shift/disbalance in covariates (participant data 
including age, sex, and so on, which influence the scans) across sites and shared support is partial (``age'' distribution in Fig. \ref{fig:overview}), 
the harmonization task becomes more involved. 
The goal is to learn 
representations as if 
the covariates were matched across the sites to begin with, see Fig. \ref{fig:overview}.
Note that in contrast to covariate shift 
methods \citep{JMLR:v10:bickel09a}, the shift here is not between train and test distributions, rather between different sources of data in the training set itself.

One recent attempt in \citep{vishnu22} to learn deep representations from pooled data which 
can also handle (shift+imbalance) in covariates needs a two stage pipeline. The first stage obtains latent representations of the scan which is {\em equivariant} to age. The second stage 
enforces {\em invariance} to scanner and
trained after freezing the 
first stage. Two covariates can be handled but an extension 
to {\em many} continuous (and categorical) covariates 
will need a complicated multi-stage model. 

{\bf Main Ideas.} In most learning tasks involving an equivariance or invariance criteria, the 
goal is to allow the model to benefit 
from structure and symmetry -- in 
the data, the parameter space, or both. Frequently, 
one formalizes these  
ideas using group theory seeking invariance/equivariance to the action of the group (e.g., $\mathbb{SO}(n)$, $\mathbb{SE}(n)$, $\mathbb{S}_n$) \citep{BloemReddy2020ProbabilisticSA,Kondor2018OnTG}. 
More general criteria (beyond properties native to the chosen group) requires specialized treatment. Our {\bf starting point} is based on the observation that Category theory \citep{fong_spivak_2019,MacLane_2014} provides 
a rich set of 
tools by which the ``structure'' (either among the 
covariates or the data more generally) can be expressed easily. 
It is known that equivariance, invariance and many group-theoretic constructs will emerge as special cases because category theory provides a more abstract treatment. Once our criteria are expressed in this way, during training,  
the necessary constraints on the latent space 
fall out directly,   
and the formulation gracefully handles heterogeneity in many continuous/categorical covariates.  
No ad-hoc adjustments are needed.

{\bf Contributions.}
{\bf (a)} On the \textbf{technical} side, we provide a 
general framework for imposing structure on the latent space 
by applying ideas from Category theory. Equivariant (and invariant) 
representation learning emerge as special cases. 
Further, this style of formalism 
unifies different formulations in vision 
and machine learning, under the same umbrella. 
{\bf (b)} On the \textbf{practical} side, we strictly generalize 
existing formulations to pool/harmonize multi-site datasets. 
While the existing two-stage formulation can deal with 
one categorical and one continuous covariate, our formulation 
places no restriction at all on the number of covariates. 
We show how the same model can also be used to heuristically approach certain hypothetical ``what if'' questions and 
offers competitive performance on public brain imaging 
datasets, with strictly more functionality/flexibility. 
\vspace{-0.1in}

\section{A brief review of Category Theory}\label{sec:cat_theory}

\vspace{-0.1in}
Category theory offers a way to study abstract ``structures'' \citep{maclane_45,MacLane_2014}. A {\bf Category} consists of two components \begin{inparaenum}[\bfseries (a)]
\item \textbf{Objects} that correspond to individual entities (e.g., scans/images), and \item \textbf{Morphisms}, identified by the paths (or ``arrows'') between the objects. Each Object has an identity Morphism (a self-loop is often omitted when drawing the diagram). \end{inparaenum}

\vspace{-5px}
\begin{remark}
    The composition of two Morphisms $f:S_1\rightarrow S_2$ and $g:S_2\rightarrow S_3$ is a well-defined Morphism and it is denoted as $g \circ f:S_1\rightarrow S_3$ (we should read it as $g$ after $f$). 
\end{remark}

\vspace{-5px}
\begin{example}
    Consider the Category of Sets, in which the Objects are sets and the Morphisms correspond to functions (or matchings) between sets. Since $\exists f:S_1\rightarrow S_2,\ g:S_2\rightarrow S_3$ then $\exists h=g\circ f:S_1\rightarrow S_3$.

    In the Category of Sets $id_S:S\rightarrow S$ corresponds to the identity function: $id_S(s)=s,\ \forall s\in S$.
\end{example}

\vspace{-5px}
\begin{definition}
    Functors $F:\mathcal{S}\rightarrow \mathcal{T}$ give a relationship between the Objects and the Morphisms of two different Categories: source Category $\mathcal{S}$ and target Category $\mathcal{T}$. The conditions below hold,
    
    \begin{compactenum}[\bfseries (i)]
    \vspace{-1pt}
    \begin{minipage}{0.4\textwidth}
        \setlength{\itemsep}{5pt}
        \setlength{\parskip}{0pt}
        \item $F(id_S)=id_{F(S)}\quad\forall S\in\mathcal{S}$ 
        \item $F(g\circ f) = F(g)\circ F(f)\\ \forall f:S_1\rightarrow S_2, g:S_2\rightarrow S_3$, in $\mathcal{S}$ 
    \end{minipage}
    \begin{minipage}{0.5\textwidth}
    
    \begin{tikzcd}[row sep=13px]
    S_1 \arrow[color=simple, loop left, "id_{S_1}"]\arrow[color=functor, line width=0.8, d, "F"] \arrow[color=simple, line width=0.6, r, "f"'] \arrow[color=comp, line width=0.6, rr, bend left=20, "g\circ f"] & S_2 \arrow[color=functor, line width=0.8, d, "F"] \arrow[color=simple, line width=0.6, r, "g"'] & S_3 \arrow[color=functor, line width=0.8, d, "F"] \\
    F(S_1) \arrow[color=simple, loop left, "id_{F(S_1)}"] \arrow[color=simple, line width=0.6, r, "F(f)"]  \arrow[color=comp, line width=0.6, rr, bend right=20, "F(g)\circ F(f)"'] & F(S_2) \arrow[color=simple, line width=0.6, r, "F(g)"] & F(S_3) 
    \end{tikzcd}
    \vspace{-10px}
    \end{minipage}
    \end{compactenum}  
\end{definition}

\begin{remark}
The reader will see why the specification above is general, but 
useful. The functor $F$ takes us from the first Category 
to the next in a way that the structure of the source Category (objects {\em and} arrows) is fully preserved in the target Category because of the two conditions above. 
\end{remark}

\section{Enforcing Symmetry and Structure}\label{sec:equiv_inv_comp}

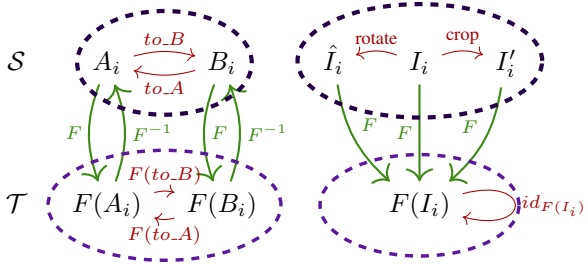
\begin{wrapfigure}{r}{0.54\textwidth}
\centering
\begin{tikzcd}[column sep=8px,every matrix/.append style = {name=m},
               remember picture,]
\mathcal{S} & A_i \arrow[color=functor, line width=0.8, dd, bend right=20, "F"'] \arrow[color=simple, r, bend left=15, "to\_B"]  
&  B_i \arrow[color=functor, line width=0.8, dd, bend right=20, "F"] \arrow[color=simple, l, bend left=15, "to\_A"] 
& & \hat{I_i} \arrow[color=functor, line width=0.8, ddr, bend right=20, "F"]  
& I_i \arrow[color=functor, line width=0.8, dd, "F"'] \arrow[color=simple, l, bend right=20, "\text{rotate}"']  \arrow[color=simple, r, bend left=20, "\text{crop}"]  
& I_i' \arrow[color=functor, line width=0.8, ddl, bend left=20, "F"']  \\ \\
\mathcal{T} & F(A_i) \arrow[color=functor, line width=0.8, uu, bend right=20, "F^{-1}"'] \arrow[color=simple, r, bend left=15, "F(to\_B)"]  
& F(B_i) \arrow[color=simple, l, bend left=15, "F(to\_A)"] \arrow[color=functor, line width=0.8, uu, bend right=20, "F^{-1}"']  
& & & F(I_i) \arrow[color=simple, loop right, "id_{F(I_i)}"] & 
\end{tikzcd}

\begin{tikzpicture}[remember picture, 
                    overlay,
                    E/.style = {ellipse, 
                                draw=source, 
                                dashed,
                                inner xsep=-3mm,
                                inner ysep=2mm, 
                                rotate=0, 
                                fit=#1,
                                line width=1.6}
                        ]
\node[E = (m-1-2) (m-1-3)] {};
\end{tikzpicture}

\begin{tikzpicture}[remember picture, 
                    overlay,
                    E/.style = {ellipse, 
                                draw=target, 
                                dashed,
                                inner xsep=-3mm,
                                inner ysep=2mm, 
                                rotate=0, 
                                fit=#1,
                                line width=1.3}
                        ]
\node[E = (m-3-2) (m-3-3)] {};
\end{tikzpicture}

\begin{tikzpicture}[remember picture, 
                    overlay,
                    E/.style = {ellipse, 
                                draw=source, 
                                dashed,
                                inner xsep=-3mm,
                                inner ysep=1.8mm, 
                                rotate=0, 
                                fit=#1,
                                line width=1.6}
                        ]
\node[E = (m-1-5) (m-1-7)] {};
\end{tikzpicture}

\begin{tikzpicture}[remember picture, 
                    overlay,
                    E/.style = {ellipse, 
                                draw=target, 
                                dashed,
                                inner xsep=3.6mm,
                                inner ysep=1.6mm, 
                                rotate=0, 
                                fit=#1,
                                line width=1.3}
                        ]
\node[E = (m-3-6) (m-3-6)] {};
\end{tikzpicture}
\vspace{-35pt}
    \caption{\footnotesize A Category theoretic view of CycleGAN (left) and SimCLR (right)}
    \label{fig:cyclegan_simclr}
    \vspace{-10pt}
\end{wrapfigure}

Symmetry/structure in the data or parameters 
are key properties we exploit in learning tasks, e.g., 
invariance in features \citep{lowe1999object}, compressive sensing \citep{candes08} as well as DNN training \citep{sabour17,hinton21}.  
Encoding symmetry/structure can involve ideas as simple as rotations and cropping
for data augmentation \citep{shorten19,dyk01,chen2020simple} to 
more sophisticated concepts 
\citep{s.2018spherical,fuchs20}.
This criteria can also be specified on the 
latent space, e.g., via geometric/topological 
priors \citep{pmlr-v119-moor20a,Ghoshetal19} or other conditions \citep{chen19,pmlr-v139-oring21a}. From Fig. \ref{fig:overview}, 
if there is a natural structure (say, in covariate space), 
then the structure 
of the representations learned on the latent space should respect that. 
To setup this description, 
we first discuss how this perspective can relate 
distinct ideas  
under an abstract (but simple) formalism. 

\subsection{Reinterpreting CycleGAN using Category theory}
Image translation uses a set of image pairs and trains
a DNN model to map between them. CycleGAN \citep{Zhu2017} is a popular 
framework for image translation 
that removes the need for paired image pairs in the training data, and learns the map from one domain (or class) to the other. 
Denoting image domains as \textit{type $A$}  
and \textit{type $B$}, the goal in CycleGAN is 
to learn how to map a \textit{type $A$} image to 
a \textit{type $B$} image and back; which we can denote as \textit{to\_B} and \textit{to\_A}: 
type $A$ (and type $B$) can mean images of horses (and zebras). 
We will avoid defining a category for the 
distributions over the images for simplicity, 
and so a training dataset made up of paired images will suffice to 
convey the key idea. 
So, for every image of type $A$, there is a similar image of type $B$. We want to learn a mapping (Functor) to a latent space that preserves the action of change from type $A$ to $B$ and back. This is shown in Fig. \ref{fig:cyclegan_simclr} (left), from which we can directly read off the following constraints,
\begin{compactenum}
    \setlength{\itemsep}{0pt}
  \setlength{\parskip}{0pt}
    \item $F$ is a fully-faithful Functor meaning:   
        $\exists F^{-1}:\mathcal{T}\rightarrow\mathcal{S} \text{ such that } F^{-1}\circ F(s) = id_s$ 
    \item Composition:  
        $\text{to\_B}\circ\text{to\_A}(B_i) =   id_{B_i}(B_i)=B_i \quad
        \text{to\_A}\circ\text{to\_B}(A_i) = id_{A_i}(A_i)=A_i$
\end{compactenum}

The first pair of constraints is 
implicit in autoencoder-like models. The second pair of constraints define the cycle consistency conditions in CycleGAN. We note that a different formalism for CycleGAN 
was described in \citep{Gavranovi__2020} by considering 
CycleGAN as a schema where cycle-consistencies enforce composition invariants in that Category.

\subsection{Reinterpreting SimCLR using Category theory}
SimCLR \citep{chen2020simple} is a \textit{self-supervised learning} method  where one attempts to map the images to a 
latent space which is invariant to transformations such as rotations, croppings, and so on. 
This can be instantiated using an invariant Functor (see Fig. \ref{fig:cyclegan_simclr}). Unlike CycleGAN, such an invariant Functor $F$, is no longer fully-faithful (i.e., $\nexists F^{-1}$ such that $F\circ F^{-1}=id_x$).

\subsection{Reinterpreting other formulations using Category theory}

{\bf (a) Latent space interpolation.} A number of approaches 
seek to perform interpolation in latent space \citep{chen19,pmlr-v139-oring21a}), informed by ``covariates'' associated with the original images (e.g., age, sex, hair length). The recipe described in the two 
examples above lends itself directly to describing such formulations, and will be subsumed by our formulation in the next section. {\bf (b) Rotation and Permutation synchronization.}
Rotation synchronization is a central problem in Cryo-EM image 
reconstruction \citep{marshall22,bendory22,tagare08}. In computer vision, 
permutation synchronization \cite{pachauri2013solving} seeks to find 
replicable matches of keypoints across many images of the same 
3D scene, e.g., for 3D reconstruction from datasets 
such as PhotoTourism \citep{li22,QuantumSync2021,birdal2019probabilistic} as well as in robotics \citep{leonardos17}. 
Since the synchronization task always involves 
consistency constraints over the symmetric Group, 
and Groups are a Category, the reinterpretation is immediate. 
{\bf (c) Equivariance and Invariance.}
Many problems in learning benefit from equivariance 
and invariance. Recall that equivariance is formally defined in 
terms of an action of a Group $G$.

\vspace{-5px}
\begin{definition}[Equivariance]
A mapping $f : \mathcal{S} \rightarrow \mathcal{T}$ defined over measurable Borel spaces $\mathcal{S}$ and $\mathcal{T}$ is said to be $G$-equivariant under the action of group $G$ iff
\begin{equation}
    f (g \cdot s) = g \cdot f (s),\quad g \in G
\end{equation}
\end{definition}
\vspace{-5px}
A more general definition of equivariance follows directly from the Functor's definition.

\begin{definition}
    A Functor $F:\mathcal{S}\rightarrow\mathcal{T}$ is defined as a mapping from $\mathcal{S}$ to $\mathcal{T}$ such that
    \vspace{-5px}
    \begin{align}
        F(id_{S}) &= id_{F(S)}\quad\forall S\in\mathcal{S} \\
        F\big(g(S_1)\big) &= F(g)\big(F(S_1)\big),\quad\forall s, g:S_1\rightarrow S_2 \in\mathcal{S}
    \end{align}
\end{definition}
\vspace{-5px}
If $\mathcal{S},\mathcal{T}$ are Borel spaces, and $g, F(g)$ belong to a Group $G$, then we obtain the group-theoretic equivariance definition. Category theory gives a more general result with no restrictions on the form of the relationships (or Morphisms) $g$.
Similarly, 
invariance (e.g., in Group theory) is defined as

\vspace{-5px}
\begin{definition}
A mapping $f : \mathcal{S} \rightarrow \mathcal{T}$ defined over measurable Borel spaces $\mathcal{S}$ and $\mathcal{T}$ is said to be $G$-invariant under the action of group $G$ iff
\begin{equation}
\vspace{-5px}
    f (g \cdot s) = f (s),\quad g \in G
\end{equation}
\end{definition}
In Category theory, this results in a special type of Functor:

\vspace{-5px}
\begin{definition}[Invariance]
A Functor $F:\mathcal{S}\rightarrow\mathcal{T}$ is invariant to the Morphism $g:S_1\rightarrow S_2$ if $F(g)=id_{F(S)}$.
\end{definition}

\subsection{A simple sanity check of benefits on MNIST Dataset}
Consider the case where the objects of the source Category $\mathcal{S}$ consist of MNIST images \citep{LeCun1998GradientbasedLA}. Since these 
images represent integers on the number line, we can ask 
whether we can learn a latent space which allows algebraic manipulations,  
with one (or more) 
basic operations defined on it. If our operation of interest was ``addition'', a 
primitive we will need for counting would be the ``$+1$'' operation. 
To keep things simple, we will focus on this operation -- 
where the source Category $\mathcal{S}$ consists of Morphisms that represent the ``$+1$'' difference between two subsequent digits. 
It is easy to check that we have indeed defined 
a Category since each Object has an identity Morphism and the Morphisms compose. 
The target Category $\mathcal{T}$ (i.e., our latent space) 
can consist of vectors $v\in\mathbb{R}^n$ as objects. Orthogonal linear mappings $W\in\mathbb{R}^{n\times n}$ as Morphisms will suffice. In our pooling/harmonization problem, the morphisms will reflect traversing the axis of each covariate.

\begin{wrapfigure}{r}{0.44\textwidth}
    \centering
    \vspace{-18pt}
    \includegraphics[width=0.43\textwidth]{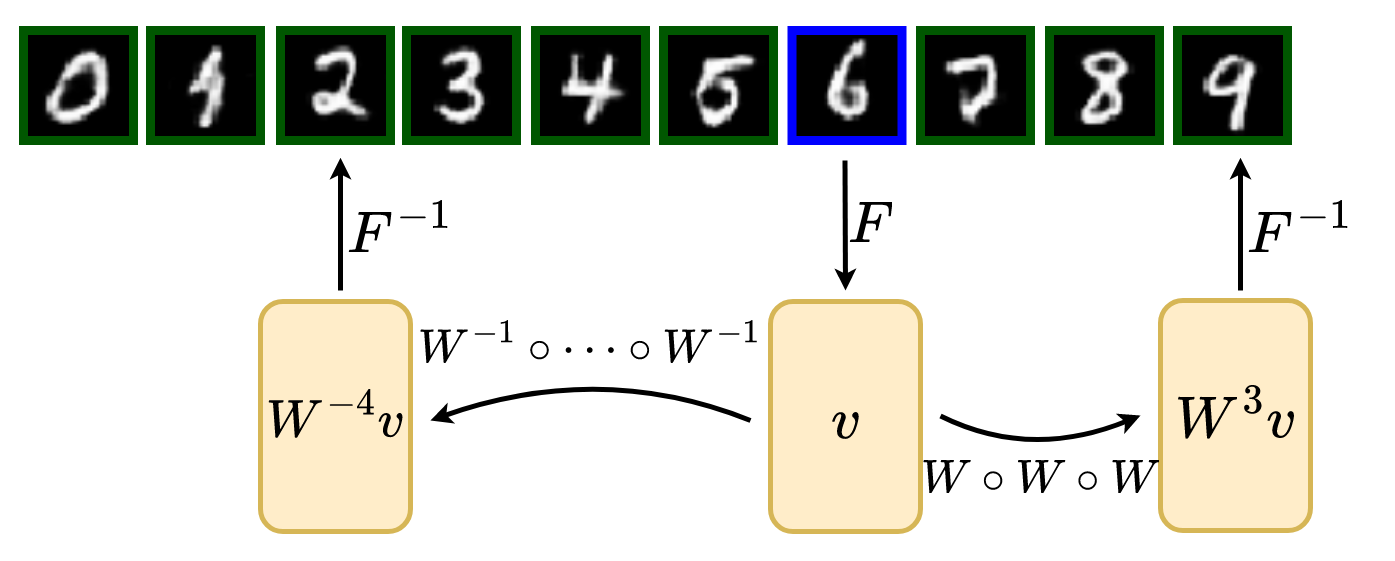}
    \caption{\footnotesize \textbf{MNIST Example:} Modelling the relationships between the digits as linear mappings in the target Category.}
    \label{fig:mnist_relationships}
    \vspace{-10px}
\end{wrapfigure}

For this MNIST example, our goal is to impose the ``$+1$'' operation in 
the latent space, see Fig. \ref{fig:mnist_relationships}. 
We define our loss to express precisely the equivalence for subsequent 
digits as well as a subset of their $+1$ compositions that can be read off of Fig. \ref{fig:mnist_relationships}. The model is {\em not} presented data for ``$-1$'' operations 
but discovers 
it due to the special structure of $W$ (orthogonal, so $W^{-1}$ exists).
Then, after training, when presented a ``seed'' image of $6$,  
the model can (forward/backward) traverse the latent space:  
a query ``$6-4$'' starts from the latent code for $6$, hops backwards 
$4$ times and generates an image of $2$. All images except ``$6$''
shown in Fig. \ref{fig:mnist_relationships} were generated in this way. 
It is worth making a brief comment on the generality here. If we had $k$ different 
morphisms, say, ``$+1$'' as well as rotation, scaling and shear, we 
could apply them in sequence and generate the corresponding images, even 
if such training data were not shown to the model (more experiments can be found in \S\ref{sec:mnist}). 
This feature 
will allow dealing with multiple covariates easily in 
our experiments. A similar solution has been provided in \cite{hae}, although their experiments were restricted to simple datasets and dealt only with 2D object dispositions. Here, we already showed how to learn something more complicated (addition) and, in the next section, we will show how to adapt this idea to complicated covariates.

\section{A Category Theory inspired Formulation for Data pooling}\label{sec:practice}

\begin{wrapfigure}{r}{0.42\textwidth}
\centering
\vspace{-20px}
\begin{tikzcd}[column sep=15px, every matrix/.append style = {name=m},
               remember picture,]
\mathcal{S} & & \mathcal{T} & Free(\mathbb{N}) \\
S_1 \arrow[color=functor, line width=0.8, rr, bend right=15, "F"] \arrow[color=simple, d, bend right=15, "f"] & & F(S_1) \arrow[color=functor, line width=0.8, r, "C"'] \arrow[color=functor, line width=0.8, ll, bend right=15, "F^{-1}"'] \arrow[color=simple, d, bend right=15, "F(f)=W"] & c_1 \\
S_2 \arrow[color=functor, line width=0.8, rr, bend right=20, "F"] 
& & F(S_2) \arrow[color=functor, line width=0.8, ll, bend right=15, "F^{-1}"'] \arrow[color=functor, line width=0.8, r, "C"']  
& c2 
\end{tikzcd}
\begin{tikzpicture}[remember picture, 
                    overlay,
                    E/.style = {ellipse, 
                                draw=source, 
                                dashed,
                                inner xsep=1mm,
                                inner ysep=2mm, 
                                rotate=0, 
                                fit=#1,
                                line width=1.6}
                        ]
\node[E = (m-2-1) (m-3-1)] {};
\end{tikzpicture}

\begin{tikzpicture}[remember picture, 
                    overlay,
                    E/.style = {ellipse, 
                                draw=target, 
                                dashed,
                                inner xsep=1mm,
                                inner ysep=1mm, 
                                rotate=0, 
                                fit=#1,
                                line width=1.3}
                        ]
\node[E = (m-2-3) (m-3-3)] {};
\end{tikzpicture}

\begin{tikzpicture}[remember picture, 
                    overlay,
                    E/.style = {ellipse, 
                                draw=free, 
                                dashed,
                                inner xsep=1mm,
                                inner ysep=2mm, 
                                rotate=0, 
                                fit=#1,
                                line width=1}
                        ]
\node[E = (m-2-4) (m-3-4)] {};
\end{tikzpicture}
\vspace{-5px}
\caption{\footnotesize A diagrammatic representation of equivariance with respect to a single covariate whose change corresponds to ``$f$'' in the original data space. Our goal is to preserve this 
structure in the latent space (Category $\mathcal{T}$) in which we model $F(f)$ as a linear transformation $W\in\mathbb{R}^{n\times n}$. In many practical cases the downstream goal is to classify the latent representations, which we formulate with the Functor $C:\mathcal{T}\rightarrow Free(\mathbb{N})$ (or $Free(\mathbb{R})$ for a regression task, etc.)}
\label{fig:equivariance}
\end{wrapfigure}
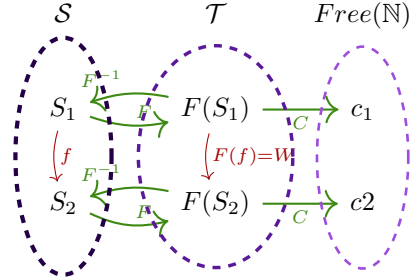

\textbf{Overview.} The preceding section provides us all the necessary modules. 
Our task involves using 
brain MR images to predict diseased or healthy controls status. 
Our covariates will include secondary (non-imaging) data 
pertaining to the participants, 
including scanner type, site, age, sex, and genotype status. 
Some of these are ordinal/continuous such as genotype (APOE; three risk types) 
and age, 
whereas others are categorical such as 
scanner type, site and sex. For the categorical covariates, 
our formulation will seek to enforce {\em invariance}
on the learned representations, i.e., asking 
that the latent representations be devoid of information 
pertinent to nuisance variables like scanner and site. 
It turns out that sex is associated 
with Alzheimer's disease (AD) because 
two-thirds of those diagnosed are women \citep{mielke18}. 
But if our goal is to understand which image features are 
relevant for the disease (and not simply to maximize accuracy), it makes 
sense to control for sex as a nuisance variable and add it 
separately at the last layer, if needed. For 
ordinal/continuous 
covariates, we see both shifts and imbalances across sites or 
scanner (different sites/scanners
may not cover exactly the same age range of participants or include
the same number of individuals for each genotype value).  
Equivariance will allow adjusting for these shifts, i.e., a coordinate system where 
the latent representations are ``aligned'' (modulo their covariate 
induced morphism). The diagram in Fig. \ref{fig:equivariance}
will setup our overall formulation, and we use the example below to describe its
operation.

\vspace{-5px}
\begin{example}
Assume we want to control only for age when learning the latent representations of the images for 
objects/participants $S_1$ and $S_2$. 
$S_1, S_2$ are identical for all covariates but have different ages. Their scans are also different. 
If $S_2$ is $x$ years older than $S_1$, in $\mathcal{S}$, we have $S_2\cong f_x(S_1)$ (i.e., equal up to isomorphism. In practice, isomorphism can be as strict as $\ell_2$-norm, cosine similarity as in CLIP-based models \citep{clip}, or even distribution-based measures such as MMD). For someone, 
$2x$ years older, similar to Fig. \ref{fig:mnist_relationships}, we will compose $f_x$ twice. 
In $\mathcal{T}$, latent representations $T_1, T_2$ correspond to $F(S_1)$ and $F(S_2)$. 
We seek to learn a functor $F$ by learning $g_x \equiv F(f_x)$ such that 
$T_2\cong g_x(T_1)$. 
If $S_1$ and $S_2$ differed in two covariates, by amounts $x$ and $y$ resp., the morphism 
from $S_1$ to $S_2$ would involve composing $f_x$ and $h_y$ (if $h$ denoted the morphisms 
for the second covariate). The morphisms in the Category $\mathcal{T}$ would compose 
similarly. 
\end{example}

\textbf{Setting up a loss function.} We will use simple morphisms in $\mathcal{T}$ which correspond to linear transformations $W\in\mathbb{R}^{n\times n}$ (this is a design choice). The pair of Functors $(F, F^{-1})$ correspond to an autoencoder, while the pair $(F, C)$ corresponds to an appropriate classifier (or regressor). 
Simply walking the paths from Fig. \ref{fig:equivariance} provides all necessary constraints 
which we write as distinct terms in our loss function, 
\begin{asparaenum}[\bfseries (i)]
    \setlength{\itemsep}{0pt}
    \setlength{\parskip}{0pt}
    \item $\small \big(F^{-1}\circ F\big)(x)=id_x$ gives the reconstruction loss:
    \vspace{-5px}
    \begin{equation}
    \vspace{-5px}
    \small
        \mathcal{L}_{r}=\sum_{s\in \mathcal{S}}\Big\|s - \big(F^{-1}\circ F\big)(s)\Big\|_2^2
    \end{equation}
    \item the prediction loss (given labels $\mathbf{y}$):
    \vspace{-5px}
    \begin{equation}
    \vspace{-5px}
    \small
        \mathcal{L}_{p} = \sum_{(i, s)\in\mathcal{S}}\underbrace{\mathcal{CE}\Big(\mathbf{y}_i, \big(C\circ F\big)(s)\Big)}_{\text{cross-entropy}}
    \end{equation}
    \item Finally, preserving the morphisms, in its simplest form, is a structure-preserving loss,
    \vspace{-5px}
    \begin{equation}
    \vspace{-5px}
    \small
        \mathcal{L}_{s} = \sum_{s_1\in S_1,s_2\in S_2}\Big\|W\cdot F(s_1)-F(s_2)\Big\|_2^2
    \end{equation}
\end{asparaenum}
The training for multiple covariates (Alg. \ref{alg:method}) is a direct generalization of the above formulation.

\vspace{-7px}
\begin{algorithm}
{\small
\DontPrintSemicolon
  
    \tikzmk{A}\KwInput{Parameters: $(\mathbf{\theta}, \mathbf{W})$, Multipliers: $\mathbf{\lambda}$}
    \KwData{input/output: $(S, \mathbf{y})\in(\mathbb{R}^{m\times n}, \mathbb{N}^m)$, and covariates: $C\in\mathbb{N}^{m\times c}$}\tikzmk{B}\boxit{input}
    \SetKwFunction{Fpairs}{pairs}
    \tikzmk{A}\For{ep in epochs}{
        $\mathcal{L}_{r}=\sum_{s\in S}\big\|s - \big(F^{-1}_{\theta_1}\circ F_{\theta_2}\big)(s)\big\|_2^2$,   $\mathcal{L}_{p} = \sum_{i=1}^m\mathcal{CE}\big(\mathbf{y}_i, \big(C_{\theta_3}\circ F_{\theta_1}\big)(S_i)\big)$, $\mathcal{L}_{s} = 0$
        
        \For {$\mathbf{c}\in C$}{
            \For {$(s_1,s_2), d$ in pairs($S, \mathbf{c}$)}{
                $\mathcal{L}_{s} \mathrel{+}= \big\|W_\mathbf{c}^d\cdot F_{\theta_1}(s_1)-F_{\theta_1}(s_2)\big\|_2^2$
            }
        }
        
        $\mathcal{L} =\lambda_1\mathcal{L}_{r} + \lambda_2\mathcal{L}_p + \lambda_3\mathcal{L}_s$
        
        step$\big(\mathcal{L},(\theta_1, \theta_2, \theta_3, W_1, ..., W_c)\big)$\tcp*{update}
    }\tikzmk{B} \boxit{train}
    \tikzmk{A}\SetKwProg{Fn}{Function}{:}{}
    \Fn{\Fpairs{$S$, $\mathbf{c}$}}{
        pairs = []
        
        \For{(i, j) in $m\times m$}{
            
                \If{$S_i$ is paired with $S_j$}{
                    pairs.append$\big((S_i, S_j), \mathbf{c}_i-\mathbf{c}_j\big)$
                }
            
        }
        \KwRet{pairs}
    }\tikzmk{B}\boxit{function}
\caption{Structure preserving training of Functors}
\label{alg:method}
}
\end{algorithm}
\vspace{-10px}

Enriching the latent space with structure provides some information 
about ``hypotheticals''. Given that we are unlikely to be provided 
different versions of a sample (one for each composition of the known Morphisms), 
a structure-preserving latent space gives us a way to: 
\begin{asparaenum}[\bfseries (i)]
    \setlength{\itemsep}{0pt}
    \setlength{\parskip}{0pt}
    \item generate new data samples, by evaluating the expression $\small \big(F^{-1}\circ F(f)\circ F\big)(S)$.
    \item answer questions, by evaluating the expression $\small \big(C\circ F(f)\circ F\big)(S)$.
\end{asparaenum}

\textbf{Role of Category Theory.} There is a clear advantage of expressing the problem in Category Theory. It enables a more concise and straightforward representation of the problem (sometimes, apparent in hindsight), as well as additional capabilities compared to earlier works \citep{vishnu22}.



\section{Experimental evaluations}\label{sec:expr}

\textbf{Rationale.} Our motivation stems from pooling brain imaging datasets, particularly in scenarios where covariate distributions vary across sites. In the ADNI brain imaging study \citep{mueller05}, while acquisition protocols are consistent among 50+ sites, demographic distributions differ. The diversity of scanners across sites, along with the impact of scanner upgrades on analysis tasks \citep{chen2020removal,ashford21,lee2019estimating}, further complicates the scenario. Scanner variations, even within the same study (e.g., ADNI-3), introduce controlled nuisances for standard analyses. Beyond ADNI, we also perform analysis on the ADCP dataset \cite{vishnu22}. Our objective is to incorporate this functionality into representation learning for regression/classification.

This problem setting underscores limitations in existing methods. Recent VAE or GAN-based approaches \citep{segmentation, deepcombat, moyer20, stargan} target scanner invariance, often accommodating binary or multiple scanner variables. However, handling variations in other covariates (e.g., age, sex, APOE) remains challenging \citep{husain21} (see \S\ref{sec:datasets}).

\textbf{Baselines.} The following baselines address distributional differences in the data while maintaining associations with covariates:
\begin{inparaitem}
\setlength{\itemsep}{0pt}
\setlength{\parskip}{0pt}
\item[\textbf{1.}] \textit{Naive}: Pooling data without pre/post processing.
\item[\textbf{2.}] \textit{MMD} \citep{li2014learning}: Minimizes Maximum Mean Discrepancy (MMD) for invariance but lacks equivariance.
\item[\textbf{3.}] \textit{CAI} \citep{Xie2017ControllableIT}: Achieves invariance with a discriminator but lacks equivariant mappings.
\item[\textbf{4.}] \textit{SS} \citep{pmlr-v70-zhou17c}: Divides the population into subgroups (Subsampling-SS) and minimizes MMD for each subgroup.
\item[\textbf{5.}] \textit{RM} \citep{Motiian2017UnifiedDS}: Matches similar samples from different scanners for invariance.
\item[\textbf{6.}] \textit{GE} (Group Equivariance) \citep{vishnu22}: Uses group theory to minimize MMD while seeking equivariance with respect to one covariate. GE achieves the best overall accuracy but has limitations: it uses expensive matrix exponentials and relies on a two-stage training, handling only two covariates.
\end{inparaitem}

\textbf{Evaluations.}
As the classification task, we will predict Alzheimer's disease (AD)/Control normals (CN) labels using the invariant and/or equivariant representations. We 
will use the following 
metrics to systematically assess each algorithm. \begin{inparaenum}[\bfseries (a)]
    \setlength{\itemsep}{0pt}
    \setlength{\parskip}{0pt}
    \item Accuracy ($\mathcal{ACC}$): Test set 
    accuracy of whether or not a participant has AD. Accuracy is reported in order to ensure that the model, despite its extra constraints, maintains all the required information that an input image carries.
    \item $\mathcal{MMD}$: MMD is a measure of distributional differences, defined as the Euclidean distance of the kernel embeddings means (typically an RBF kernel).
    \item $\mathcal{ADV}$: an alternative to measure invariance is by training a model to predict the nuisance covariate (e.g., ``Scanner'') using the latent representations. If we obtain latent representations 
    devoid of this information, then the accuracy should be almost random.
\end{inparaenum}
The above two metrics identify invariance with respect to scanners. To check equivariance to covariates, we use two measures: Minimum distance ($\mathcal{D}$) and Cosine similarity ($\mathcal{CS}$), 
\begin{asparaenum}[\bfseries (a)]
        \item Assume that a value of $c_1$ (for a covariate $\mathbf{c}$) corresponds to an individual $s_1$, and Morphism $W\in\mathbb{R}^{n\times n}$ in the latent space gives a change
        in the covariate from $c_1$ to $c_2$. Then, the Minimum Distance (similar to covariate matching) identifies the closest individual $s$ such that $s_{\mathbf{c}}=c_2$:
        \begin{equation}
            \small
            \mathcal{D}(s_1; c_2) = \min_{s:s_{\mathbf{c}}=c_2} \frac{\big\|W\cdot F(s_1)-F(s)\big\|_2}{n}
            \vspace{-6px}
        \end{equation}
        
        \item A low minimum distance by itself is insufficient. So, we also calculate the cosine similarity ($\mathcal{CS}$) between all the other covariates of $s_1$ and $s_2$, where
        $\small s_2 = \arg\min_{s:s_{\mathbf{c}}=c_2} \big\|W\cdot F(s_1)-F(s)\big\|_2$.
\end{asparaenum}
    Low Minimum Distance ($\mathcal{D}$) and high Cosine Similarity ($\mathcal{CS}$) is desirable.
    
\textbf{Setting.}
We model the Functor $F$ using a modified ResNet  \citep{he16}, and the Functor $C$ using a Fully-Connected Neural network. For our experiments, 
using linear mappings $W\in\mathbb{R}^{n\times n}$ 
for the Morphisms in the target Category $\mathcal{S}$ (i.e., latent space) was sufficient but this can be 
easily upgraded. 
All the experiments are a result of $5$-fold cross validation procedure.
In the following sub-sections, we summarize our experimental findings. We gradually increase the complexity of the experiments (number of constraints that we simultaneously optimize).

{\bf Can Category theory constraints help impose invariance to scanner?}
The following objective derived directly from the  formulation is used, 

\vspace{-15px}
\begin{equation}
    \small
    \mathcal{L} = \underbrace{\mathcal{L}_p}_{\text{cross-entropy}} +  \sum_{s_1, s_2\in\mathcal{S}}\underbrace{\lambda\cdot \Big\|F(s_1) - F(s_2)\Big\|}_{\forall\ s_1, s_2 \text{ from different Scanners}} 
    \label{eq:inv}
    \vspace{-10px}
\end{equation}

Our results  (Table \ref{tab:results}) show that we obtain invariant representations without degrading the model's accuracy,  consistent with the literature on invariant representation learning \citep{moyer2018invariant}. MMD decreases by more than $80\%$ compared to the best baseline, while accuracy is $1.3\%$ higher than the naive model, suggesting stronger generalization to new samples (more experiments on tabular data can be found in \S\ref{sec:tabular}, and ablation studies on $\lambda$ and latent space size in \S\ref{sec:lambda}).

\textbf{Can a model be invariant to scanner and remain equivariant to covariates?}
Besides invariance, we seek equivariance to specific covariates in some cases. Here, we examine the simplest case of equivariance to a single covariate. We test three different covariates: \begin{enumerate*}
    \item Age
    \item Sex
    \item APOE A1
\end{enumerate*}. Since age is continuous, we discretize it into bins of 10 years (so we have enough samples in each 
bin). APOE A1 (and A2) are genotypes in ADNI that are associated with AD \citep{husain21,sienski21}.
Using our formulation, we derive the following loss function,

\vspace{-15px}
\begin{equation}
    \small
    \hspace{-0.1in}\mathcal{L}= \hspace{-0.1in}\underbrace{\mathcal{L}_p}_{\text{cross-entropy}} \hspace{-0.1in} +  \sum_{s_1, s_2\in\mathcal{S}}\underbrace{\lambda\Big\|W^{c_{s_1}-c_{s_2}} \cdot F(s_1)- F(s_2)\Big\|}_{c_{s_1}, c_{s_2} \text{ represent the nuisance covariate}}
    \label{eq:equiv}
\end{equation}
\vspace{-10px}

Note that the invariance term is implicit in this case because for two individuals $s_1, s_2$ with $c_{s_1}=c_{s_2}$ we recover the invariance term from (\ref{eq:inv}).

\begin{wraptable}{r}{0.47\textwidth}
    \vspace{-15px}
    \centering
    {\scriptsize
    \begin{tabular}{l|c c c : c}
    & \multicolumn{3}{c}{Parameters} & \multirow{2}{*}{Runtime} \\
    & Encoder & Classifier & Morphisms &    \\
    \hline
        GE & $\mathcal{O}(\mathcal{E})$ & $\mathcal{O}(\mathcal{C})$ & $\mathcal{O}(\mathbf{c}\cdot \mathcal{P})$ & $\mathcal{O}(\mathbf{c}\cdot \mathcal{R})$ \\
        \textbf{Ours} & $\mathcal{O}(\mathcal{E})$ & $\mathcal{O}(\mathcal{C})$ & $\mathcal{O}(\mathbf{c})$ & $\mathcal{O}(\mathcal{R})$
    \end{tabular}
    }
    \vspace{-5px}
    \caption{\footnotesize \textbf{Parameters / Runtime comparison with respect to the number of covariates.} $\mathbf{c}$ is the number of covariates and $\mathcal{P}$ the number of unique pairs for each covariate (exponential growth). $\mathcal{E}/\mathcal{C}$ is the number of parameters in the Encoder and Classifier respectively (independent of $\mathbf{c}$). $\mathcal{R}$ stands for a single training runtime. Our method's complexity is better both in parameters and runtime.}
    \label{tab:params}
    \vspace{-10px}
\end{wraptable}

\vspace{-5px}
\begin{remark}
While this specific problem setting is not new (see \citep{Motiian2017UnifiedDS}), here we show that the simplicity of our method comes with certain \textbf{advantages}. First, we get improved results and further, our model has fewer parameters and smaller runtime complexity (Tables \ref{tab:results}, \ref{tab:params}). 
\end{remark}
\vspace{-5px}

\begin{wrapfigure}[22]{r}{0.3\textwidth}
    \centering
    \vspace{-8px}
     \includegraphics[width=0.3\textwidth]{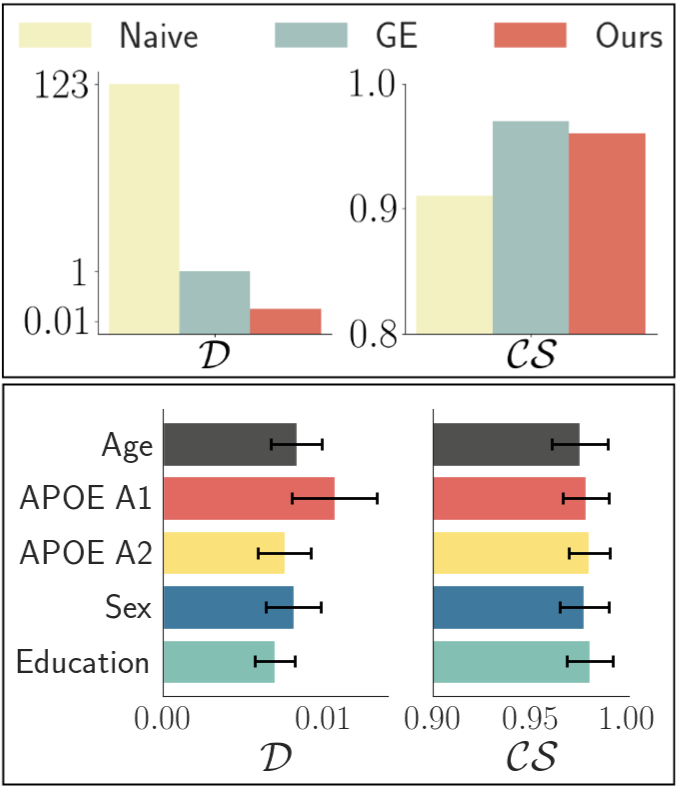}
     \vspace{-20px}
     \caption{\footnotesize \textbf{(Left)} Minimum Distance $(\mathcal{D})$ and Cosine Similarity $(\mathcal{CS})$ in for age equivariance, compared with Naive and GE. \textbf{(Right)} Minimum Distance $(\mathcal{D})$ and Cosine Similarity $(\mathcal{CS})$ for 5-covariate equivariance. Results are consistently good.}
     \label{fig:d_cs}
\vspace{-10px}     
\end{wrapfigure}

In Fig. \ref{fig:d_cs}, we examine the performance of the naive model where we infer $W$ post-training, GE \citep{vishnu22} which defines a different linear transformation $W$ for each age difference, and our model (Ours). In all three models, we model the age increase (e.g., $60$ to $65$) by applying a linear transformation in the latent space. 
Then, we find the closest point 
for that age (e.g., $65$) in the latent space
w.r.t. $\ell_2$ norm ($\mathcal{D}$) and cosine similarity ($\mathcal{CS}$). 
The transformed vector should be in the neighborhood of 
latent vectors with that age (e.g., $65$), and so $\mathcal{D}$ should be small and $\mathcal{CS}$ should be high.
The results show that such a morphism is not accurate enough in the naive or even the GE algorithm. The naive model tends to ``spread'' the latent vectors without preserving structure, so it cannot capture equivariance in the latent space 
(clear when we examine $\mathcal{D}$), while GE achieves equivariance but our results 
suggest improvements.

\begin{table}[tb]
    \centering
    {\scriptsize
    \begin{tabular}{l | c: c c || c: c c }
    & \multicolumn{3}{c||}{ADNI} & \multicolumn{3}{c}{ADCP} \\ 
     & $\mathcal{ACC}\uparrow$ & $\mathcal{MMD}(\times 10^2)\downarrow$ & $\mathcal{ADV}\downarrow$ & $\mathcal{ACC}\uparrow$ & $\mathcal{MMD}(\times 10^2)\downarrow$ & $\mathcal{ADV}\downarrow$  \\ [0.5ex] 
     \hline\hline
      Random &  $64$   &  -     &  $49$ &  $74$   &  -     &  $42$    \\
      \hline
     \textit{Naive} & $80{\scriptstyle(2.6)}$ & $27{\scriptstyle(1.6)}$ & $59{\scriptstyle(2.9)}$ & $83{\scriptstyle(4.4)}$ & $90{\scriptstyle(08.7)}$ & $49{\scriptstyle(08.4)}$\\ 
     \textit{MMD}\citep{li2014learning} & $80{\scriptstyle(2.6)}$ & $27{\scriptstyle(1.8)}$ & $59{\scriptstyle(3.3)}$ & $84{\scriptstyle(6.5)}$ & $86{\scriptstyle(11.0)}$ & $49{\scriptstyle(11.9)}$ \\
     \textit{CAI}\citep{Xie2017ControllableIT} & $74{\scriptstyle(3.6)}$ & $27{\scriptstyle(1.5)}$ & $61{\scriptstyle(2.1)}$ & $82{\scriptstyle(5.1)}$ & $85{\scriptstyle(12.3)}$ & $56{\scriptstyle(06.9)}$  \\
     \textit{SS}\citep{pmlr-v70-zhou17c} & $81{\scriptstyle(3.7)}$ & $26{\scriptstyle(1.6)}$ & $57{\scriptstyle(2.1)}$ & $82{\scriptstyle(3.5)}$ & $88{\scriptstyle(14.6)}$ & $51{\scriptstyle(06.7)}$  \\
     \textit{RM}\citep{Motiian2017UnifiedDS} & $78{\scriptstyle(3.8)}$ & $22{\scriptstyle(0.6)}$ & $52{\scriptstyle(5.4)}$ & $84{\scriptstyle(5.3)}$ & $77{\scriptstyle(13.8)}$ & ${\bf 40}{\scriptstyle(04.7)}$ \\ 
     \textit{GE}\citep{vishnu22} & $77{\scriptstyle(4.8)}$ & $16{\scriptstyle(7.2)}$ & ${\bf 50}{\scriptstyle(4.2)}$ & $81{\scriptstyle(1.8)}$ & $70{\scriptstyle(22.3)}$ & $49{\scriptstyle(07.3)}$ \\
     \hline

     \textbf{Ours} (inv) & $81{\scriptstyle(2.3)}$ & ${\bf 02}{\scriptstyle(2.4)}$ & $52{\scriptstyle(2.0)}$ & ${\bf 86}{\scriptstyle(8.0)}$ & ${\bf 34}{\scriptstyle(01.2)}$ & $45{\scriptstyle(04.2)}$  \\
     \textbf{Ours} (1 cov) & ${\bf 82}{\scriptstyle(2.5)}$ & ${\bf 11}{\scriptstyle(2.9)}$ & $53{\scriptstyle(1.2)}$ & ${\bf 86}{\scriptstyle(5.9)}$ & ${\bf 39}{\scriptstyle(03.6)}$ & $48{\scriptstyle(05.7)}$  \\
     \textbf{Ours} (2 cov) & ${\bf 82}{\scriptstyle(2.2)}$ & ${\bf 10}{\scriptstyle(6.6)}$  & $51{\scriptstyle(3.0)}$ & ${\bf 85}{\scriptstyle(7.7)}$ & ${\bf 40}{\scriptstyle(04.3)}$ & $44{\scriptstyle(08.1)}$ \\
     \textbf{Ours} (5 cov) & $80{\scriptstyle(2.3)}$ & ${\bf 11}{\scriptstyle(5.3)}$ & $52{\scriptstyle(1.7)}$ & $-$ & $-$ & $-$ \\
     \hline\hline
    \end{tabular}
    }
    \caption{\footnotesize \textbf{Quantitative Results on ADNI \citep{mueller05} and ADCP from \citep{vishnu22}.} The mean accuracy ($\mathcal{ACC}$) and invariance as evaluated by $\mathcal{MMD}$ and $\mathcal{ADV}$ are shown. The standard deviations are in parenthesis. The baseline \textit{inv} corresponds to only invariance to Scanner, \textit{$x$ cov} corresponds to equivariance with respect to $x$ covariates (along with invariance to Scanner). $\mathcal{MMD}$ measure is significantly reduced without any drop in accuracy. The $\mathcal{ADV}$ measure is close to the random baseline (desirable). }
    \label{tab:results}
    \vspace{-20px}
\end{table}

\begin{wrapfigure}{r}{0.5\textwidth}
    \centering
    \vspace{-15px}
     \includegraphics[width=0.5\textwidth]{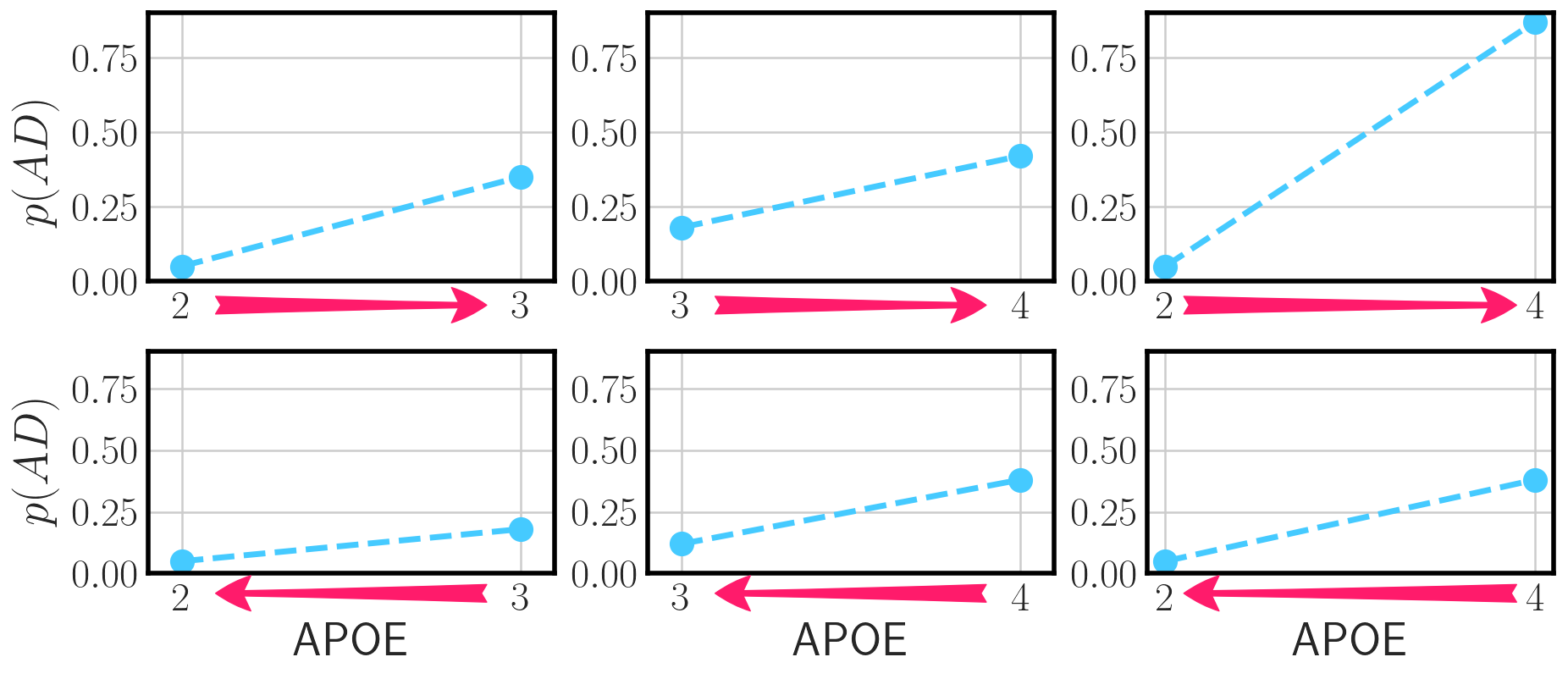}
     \vspace{-20px}
     \caption{\footnotesize Increasing APOE A1 in the latent space using morphisms leads to a higher probability of AD and vice versa. Such manipulations in latent space are feasible through learned morphisms.}
     \label{fig:apoe_interv}

\vspace{-10px}
\end{wrapfigure}

\textbf{Assessing hypothetical samples.}
Enriching the latent space with structure provides information about questions such as \textit{What would be the change in AD if APOE A1 had a different value?} Recall that APOE A1 is a ordinal covariate with 3 values ($\{2,3,4\}$) and associated with AD \citep{husain21,sienski21}. For each sub-group 
(individuals with a specific value for APOE A1), we increase/decrease its value using morphisms and then classify the new vector. An increase in APOE A1 results 
in a higher probability of AD and vice-versa (Fig. \ref{fig:apoe_interv}), as expected \citep{husain21}.

\section{Related work}\label{sec:related}

\textbf{Data pooling, Fairness, disentanglement, and Invariant Representation Learning.}
General approaches for analyzing data  
from different sites involve meta analysis \citep{enigma,rucker21}. 
When data transfer is feasible, pooling can be approached using 
the Johnson-Neyman technique (e.g., when ANCOVA is inapplicable). 
Some tools 
from statistical genomics have been deployed for brain image data pooling \citep{johnson2007adjusting}, also see \citep{GarciaDias2020NeuroharmonyAN}.
For deep models, 
the link between invariant representation learning/fairness 
and data pooling was seen in \citep{moyer2018invariant,Jaiswal2019UnifiedAI,jade}, and has been 
the defacto approach \citep{moyer20,vishnu22,Peng2017ReconstructionBasedDF,LIU2022102516,pham2023fairness} to 
disentangle the influence of nuisance variables (but provably doing so 
is difficult \citep{locatello22}).

\textbf{Category theory in machine learning.}
The applications of Category theory in 
applied disciplines are somewhat limited.
But more recently, these ideas have been successfully applied 
to other fields (e.g. \citep{fong_spivak_2019}). In machine learning specifically, \citep{Fong2019BackpropAF,Cruttwell2022CategoricalFO,Fong2019LensesAL,barbiero2023categorical} model the learning process of gradient-based learning algorithms using Lenses and \cite{gavranovic2024categorical} uses Monads.
Separately, \citep{Wilson2020ReverseDA} used 
Category theory to offer a framework for training Boolean circuits. 
The results in \citep{Gavranovi__2020} modeled CycleGAN using Functors 
and showed how the formulation can be used for inserting/deleting objects from an image. 
A summary of 
developments is in \citep{Shiebler2021CategoryTI}.

\section{Conclusions}\label{sec:conc}

Imposing structure on latent spaces 
learned by DNN models is being 
actively studied, for problems 
ranging from disentanglement to interpretability. Using 
a data pooling problem in brain imaging as a motivation, 
we discuss how category theory provides 
a precise framework for such tasks. 
Not only does such an approach significantly simplify 
recent works in the literature, but also offers 
a perspective unifying an array of standalone approaches/models. 
Due to its rather
abstract nature, the application of category theory in vision/machine learning is 
rather limited. 
Nonetheless, we believe that the models/experiments 
in this paper provide evidence that these ideas can inform other challenging 
problems in our field, including explainability, modularity and interpretability. 
 \textbf{Limitations}. We point out a few key caveats. Since the goal was to demonstrate the benefits of the formalism rather than maximize accuracy attained for each downstream task, the choice of most modules was kept quite simple (linear transformations, ResNets etc). Our construction offers 
 a great deal of flexibility to upgrade these modules, as needed. 
 However, due to the nature of our algorithm (i.e., imposing constraints on the latent space), we cannot directly benefit from contemporary architectures 
 such as U-Net \citep{unet, unet1, cyclegan1} since their skip-connections provide shortcuts that will avoid the category-theory informed constraints. This means that, in contrast to some other works that harmonize two sets of images (mostly, with one categorical 
 nuisance variable), the proposed method is best suited for obtaining a well-behaved latent space with 
 multiple nuisance variables/covariate shifts, for a broad range of downstream tasks.

\textbf{Acknowledgments}. The authors thanks Veena Nair and Vivek Prabhakaran from UW Health for help with
the ADCP dataset. Research was supported by NIH grants RF1AG059312 and RF1AG059869, as well as NSF award CCF 1918211 and the Bodosaki scholarship.

\newpage

{\small
\bibliographystyle{iclr2024_conference}
\bibliography{main}
}

\appendix

\section{A warm-up analysis}\label{sec:mnist}

\subsection{Multi-equivariance}

In this section, we evaluate the performance of our approach when we require the latent space to be equivariant with respect to two different transformations. We consider a dataset of images and the goal is to create representations that are equivariant with respect to image rotations as well as image scaling. 

We use the MNIST dataset \cite{LeCun1998GradientbasedLA} as a toy example to illustrate the key idea. 
We assume that the latent space is the Category with the following characteristics:
\begin{compactenum}[\bfseries (a)]
    \item Objects: vectors $\in\mathbb{R}^n$
    \item Morphisms: orthogonal linear transformations $W\in\mathbb{R}^{n\times n},\ W^T W= I$ (the identity morhpism is the identity matrix $I$)
\end{compactenum}

We learn an autoencoder (i.e., a pair of fully faithful Functors $F, F^{-1}$) and two matrices $W_r, W_s$ that represent the rotation and scale morhpism respectively in $\mathbb{R}^n$. In this experiment, we set $n=128$. During training, we provide our model with an image from MNIST, along with a rotated version (of the same image) and separately, a scaled one. An important note is that \textbf{we do not provide any images that are both rotated and scaled during training.} We define 
\begin{compactenum}
    \item \textit{Rotate} to be a counter-clockwise rotation by $5$ degrees. This means that a rotation by e.g. $15$ degrees corresponds to ``$\textit{Rotate}\circ\textit{Rotate}\circ\textit{Rotate}$'' in the source Category of images and to $W_r^3$ in the latent space of $\mathbb{R}^n$
    \item \textit{Scale} the operation of ``zoom-out'' in an image. To achieve this behaviour, we first pad the input image (1 extra pixel on each side) and then we resize it back to the original shape. According to this definition, the inverse operation ($\textit{Scale}^{-1}$) is defined as zooming into the image. 
\end{compactenum} 

Our training objective consists of the following loss terms:
\begin{equation}
    \mathcal{L} = \mathcal{L}_r + \lambda\mathcal{L}_s + \mu\bigg((W_r^TW - I)^2 + (W_s^TW - I)^2\bigg)
\end{equation}
The first two terms are suggested directly by our framework while the last two constraints ensure that the latent space morphisms have the desired form and are case-specific.

Although during training we provide images that are only counter-clockwise rotated (up to 50 degrees) or scaled (up to 10 paddings), our model is able to map the images to the latent space and preserve the two morphisms, which leads to the following abilities:
\begin{asparaenum}[\bfseries (a)]
    \item We can generate realistic images that are counter-clockwise rotated by more than 50 degrees by continuously applying the linear transformation $W_r$ in the latent space and mapping the new vector back to the original Category of images using the decoder ($F^{-1}$).
    \item
    We can generate realistic images that are scaled by more than 10 pads by continuously applying the linear transformation $W_s$ in the latent space and mapping the new vector back to the original Category of images using the decoder ($F^{-1}$).
    \item We can generate realistic images that are clockwise rotated by applying the inverse linear transformation $W_r^{-1}$ to the latent representation of an image and then map the new vector back to the original Category of images using the decoder ($F^{-1}$).
    
    \item We can generate realistic images that are zoomed-in by applying the inverse linear transformation $W_s^{-1}$ to the latent representation of an image and then mapping the new vector back to the original Category of images using the decoder ($F^{-1}$).
    \item Finally, we can generate realistic images that are both rotated and scaled by applying both linear transformations $W_r \circ W_s ~(=W_s \circ W_r)$ in the latent representation of an image and then mapping the new vector back to the original Category of images using the decoder ($F^{-1}$).
\end{asparaenum}

\subsubsection{Qualitative results}

\begin{figure}
     \centering
     \begin{subfigure}[b]{0.48\textwidth}
         \centering
         \includegraphics[width=\textwidth]{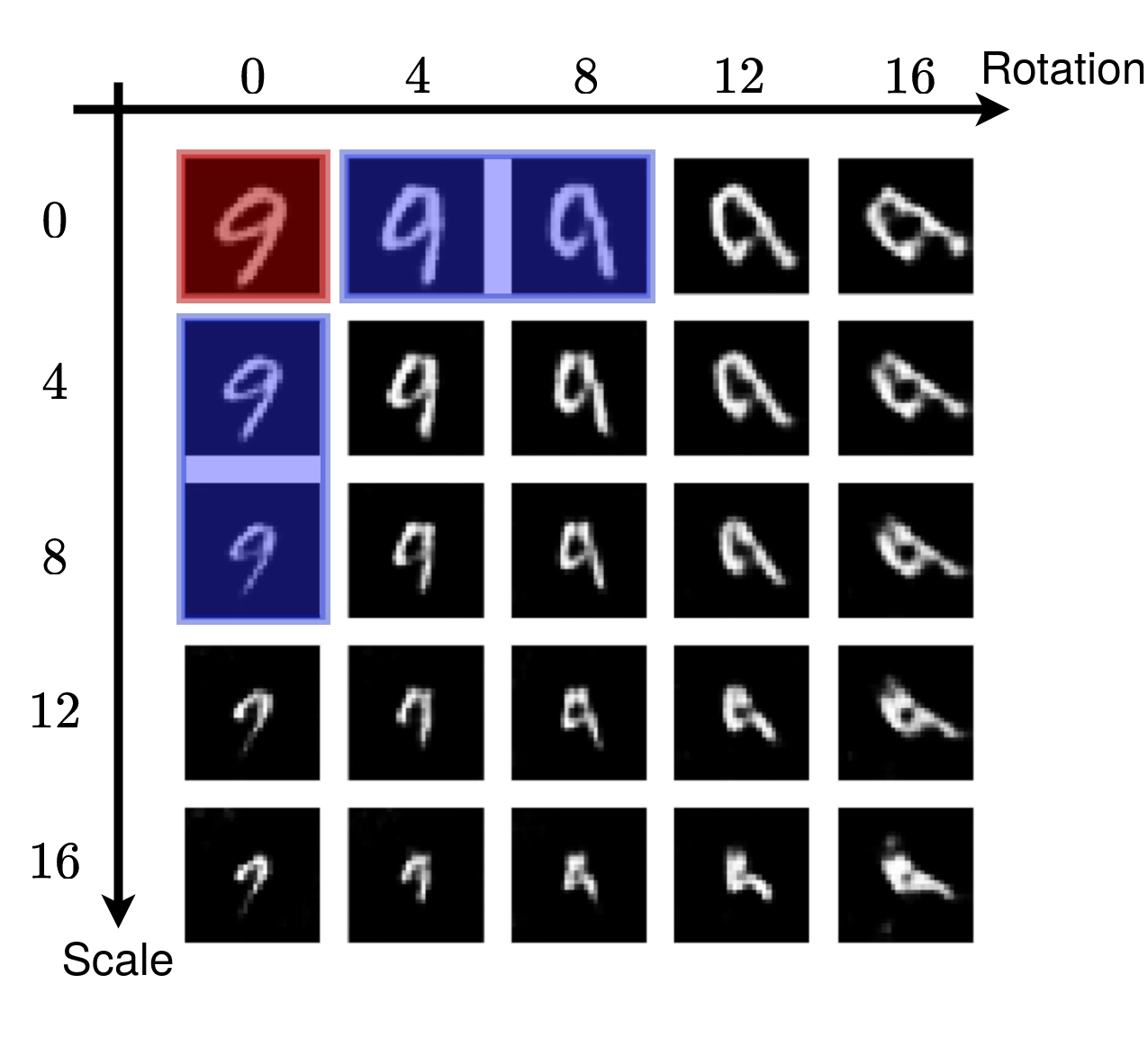}
     \end{subfigure}
     \hfill
     \begin{subfigure}[b]{0.48\textwidth}
         \centering
         \includegraphics[width=\textwidth]{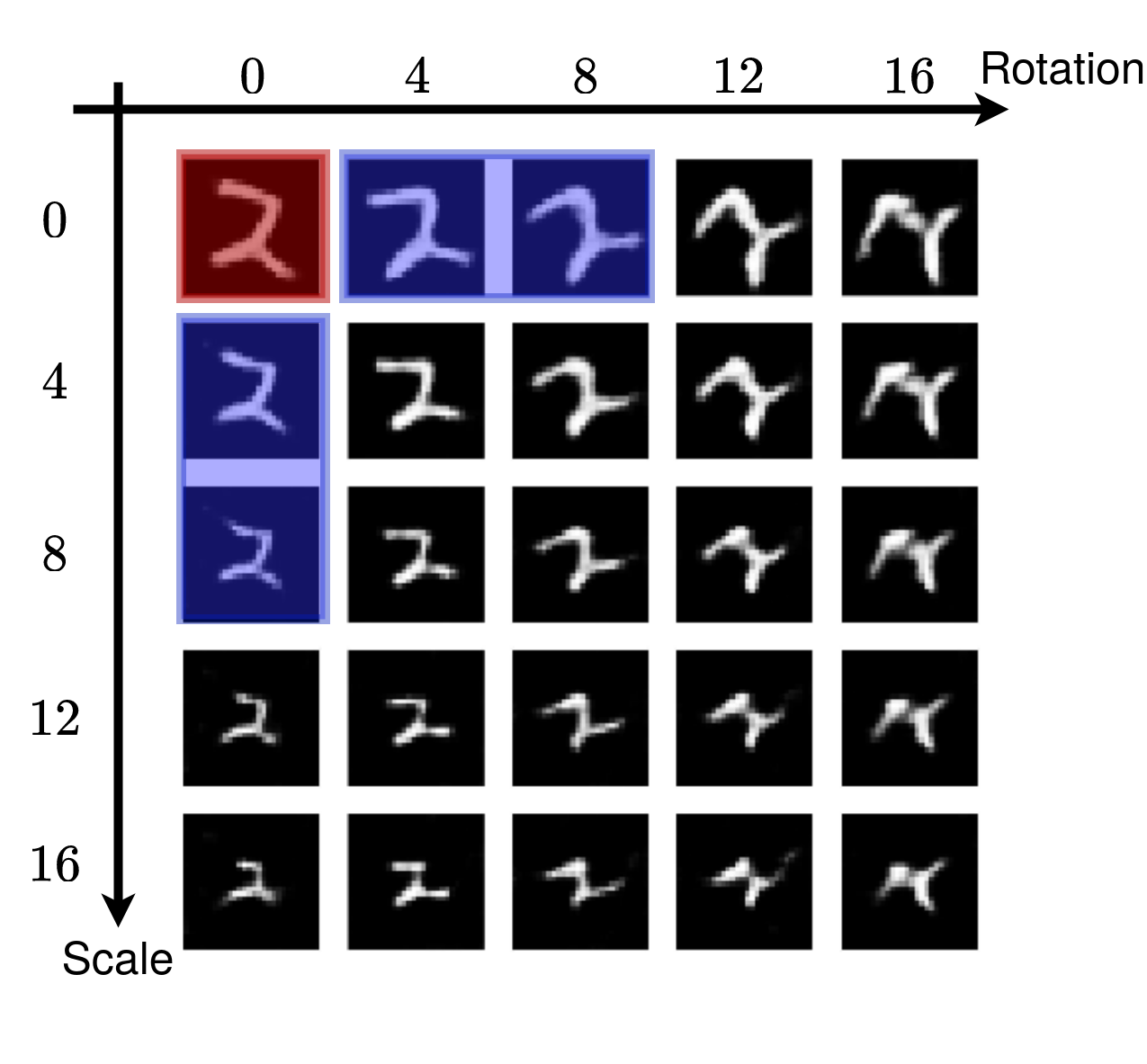}
     \end{subfigure}
     
        \caption{Composition of rotation and scaling. \textbf{The transformations were applied to the original image's latent vector ($v$) in the latent space} according to the composition rule $W_{r}^{i} W_{s}^{j} v$ where $i, j$ denote the degree of rotation and scaling respectively. In red we depict the original image, in blue we depict the only transformations that were ``visible'' during training.}
        \label{fig:comp}
\end{figure}

In Figure \ref{fig:comp}, we show the generated images when we apply both transformations to a latent vector. The 2 main takeaways from that figure are:
\begin{asparaenum}[\bfseries (a)]
    \item Our model is able to generalize beyond the presented transformations as depicted by the row/column in which scale/rotation was $0$ respectively. We observe that even after applying $W_r^{20}, W_s^{20}$ (twice as much as the data presented during training) we get realistic results that follow the rotation and scaling rules.
    \item Our model is able to compose the transformation in the latent space and generate images that follow both transformations, although no composed images were presented during training.
\end{asparaenum}

\begin{figure}
     \centering
     \begin{subfigure}{0.8\textwidth}
         \centering
         \includegraphics[width=\textwidth]{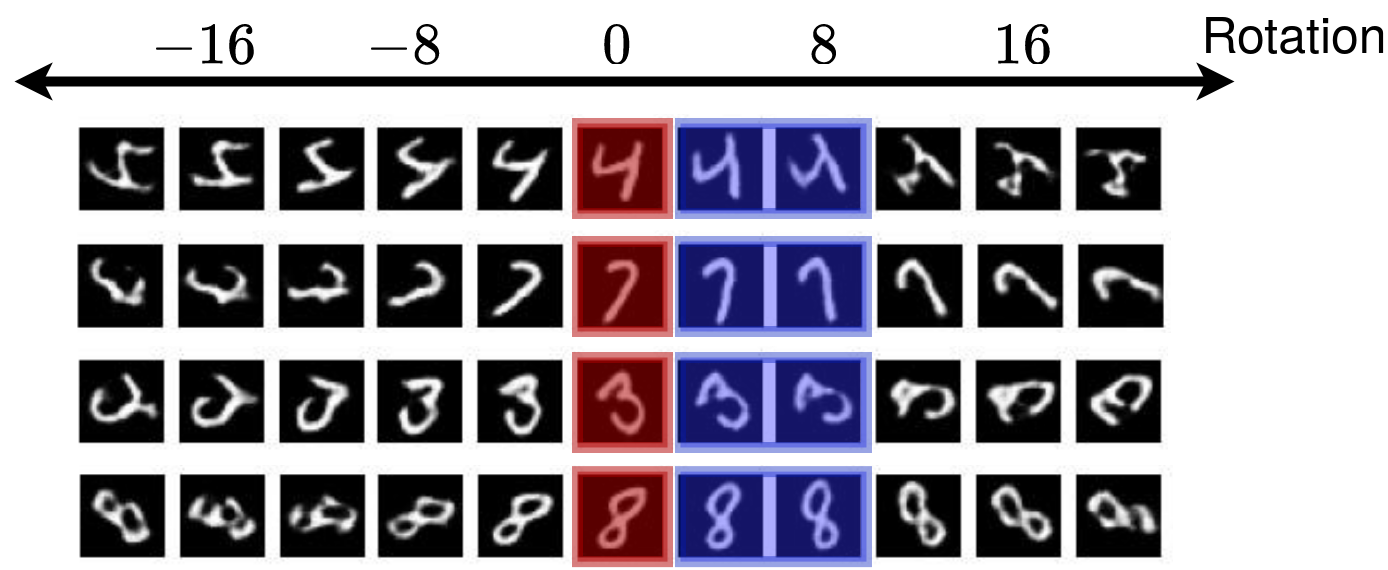}
     \end{subfigure}
     \hfill
     \begin{subfigure}{0.8\textwidth}
         \centering
         \includegraphics[width=\textwidth]{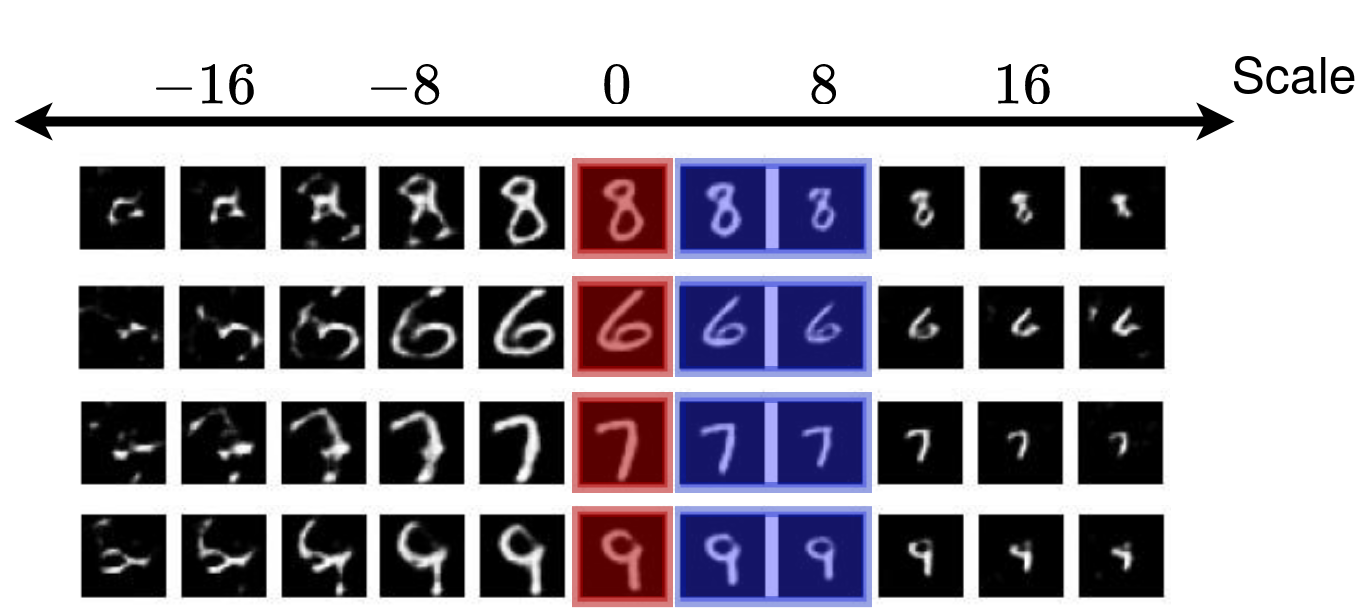}
     \end{subfigure}
     
        \caption{Rotation and Scaling in both directions. \textbf{The transformations were applied to the original image's latent vector ($v$) in the latent space} according to the morphisms $W_{r},\  W_r^{-1},\ W_s,\ W_{s}^{-1}$. In red we depict the original image, in blue we depict the only transformations that were ``visible'' during training. Inverse scaling, as expected, fails after an adequate number of steps, since the digit can no longer fit in the image. On the contrary, rotation is more robust and performs well even for large rotations in both directions.}
        \label{fig:inv}
\end{figure}

In Fig. \ref{fig:inv}, the effect of the inverse linear transformation is shown. Although there were no such images during training, our model is able to understand the two transformations due to the categorical structure imposed on the latent space. As a result, we can generate clockwise rotated and zoomed-in images with no quality drop (zoom-in fails after a significant amount of constant applications due to the space constraints while rotation remains realistic even for an extreme number of linear transformations). 

\subsubsection{Quantitative results}

\begin{figure}
    \centering
    \includegraphics[width=0.5\textwidth]{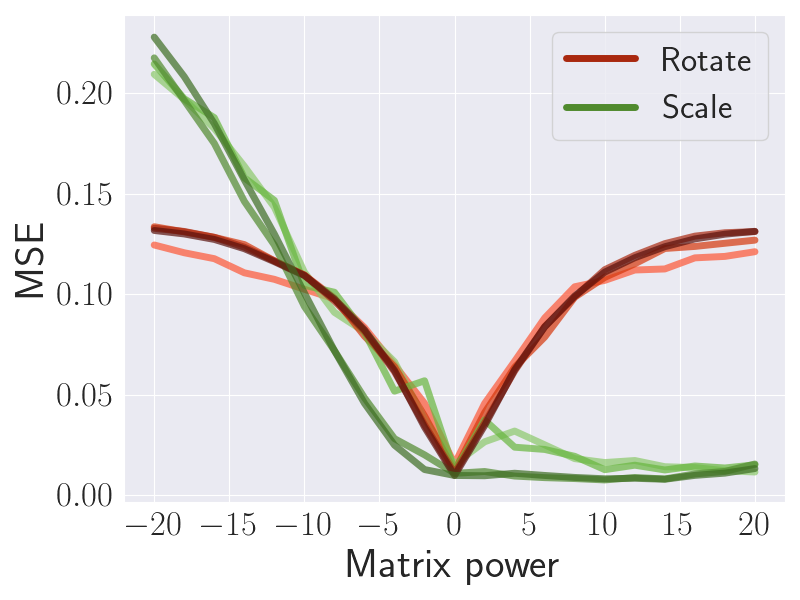}
    \caption{MSE error of the predicted image against the ground truth as we rotate it using $W_r, W_r^{-1}$ and scale it using $W_s, W_s^{-1}$. The line intensity represents the size of the latent space ($32, 64, 128, 256$) and it shows that the results are accurate enough even for small latent spaces.}
    \label{fig:rot_scale_mse}
\end{figure}

In Fig. \ref{fig:rot_scale_mse} we quantify what we observed already. Rotation provides consistently good results independent of the direction, while inverse scaling (zoom-in) fails after some steps, due to the fact that the depicted digit is too big to fit in the image. On the contrary, we can continuously apply the scale operation (zoom-out) up to the point which the digit is no larger than a couple of pixels, with no decrease in the quality. Also, the results do not vary much as we increase the latent dimension from $32$ to $256$.

\subsection{Equivariance to algebraic morphisms}\label{subsec:paired_MNIST}

Here we show that the same formulation, without any change, is able to create a mapping to the latent space which is equiavariant to a much more abstract and complex morphism (algebraic addition). Specifically, we consider the MNIST dataset and our goal is to map the images to a latent space that preserves the relationships between the digits. We use the same Category as above for our latent space and we wish to enforce the following condition: if for two images, $I_1, I_2$ with labels $l_1, l_2$ the corresponding latent representations are $v_1, v_2$, then $v_2 \simeq W^{l_2-l_1}v_1$. 

We use the same setting as in the previous experiment (but here we learn a single matrix $W$). During training, we provide our model with pairs of data that corresponds to two images $(I_1, I_2)$ with $l_1 + 1 = l_2$. This means that, during training, we do not optimize for $W^{-1}$ simultaneously (just like in the previous experiment). In this experiment we set $n=32$. 

We use the same AutoEncoder architecture and the objective in this case is:
\begin{equation}
\mathcal{L} = \mathcal{L}_r + \lambda\mathcal{L}_s + + \mu\big((W^TW - I)^2 \big)
\end{equation}
which means that we use the exact same constraints that our formulation has suggested along with the constraint that $W$ is orthogonal.

\subsubsection{Qualitative results}

\begin{figure}
    \centering
    \includegraphics[width=0.8\linewidth]{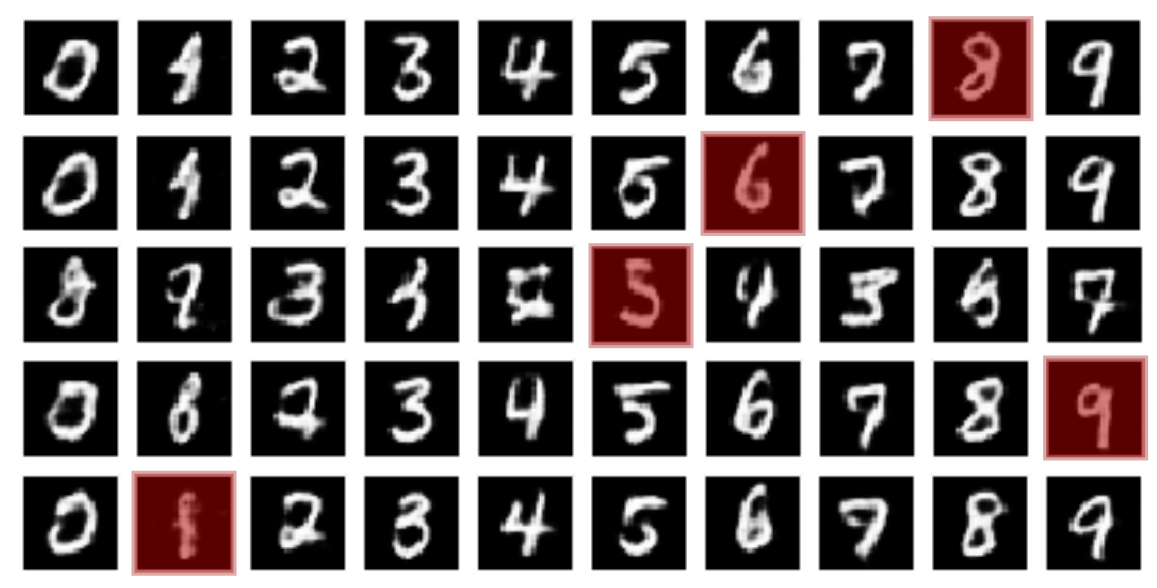}
    \caption{Images generated when we apply $W$ and $W^{-1}$ to the latent vector of the image in red. The images to the left of the red image were generated by using the linear transformation $W^{-1}$ on $v$, where $v$ is the latent vector of the original image (red), and the images to the right were generated by using the linear transformation $W$ on $v$. While there might exist some failed cases (e.g. row 3), in most cases we are able to generate realistic images for each digit. 
    }
    \label{fig:paired_digits}
\end{figure}

Figure \ref{fig:paired_digits} shows some of the generated images we obtain when we apply the linear transformations $W, W^{-1}$ in the latent vector of an image. We can observe that, altough we do not explicitly train our model for the inverse transformation ($W^{-1}$), it is able to generate realistic images due to the orthogonality constraint we impose in our model.

\begin{figure}
    \centering
    \includegraphics[width=0.8\linewidth]{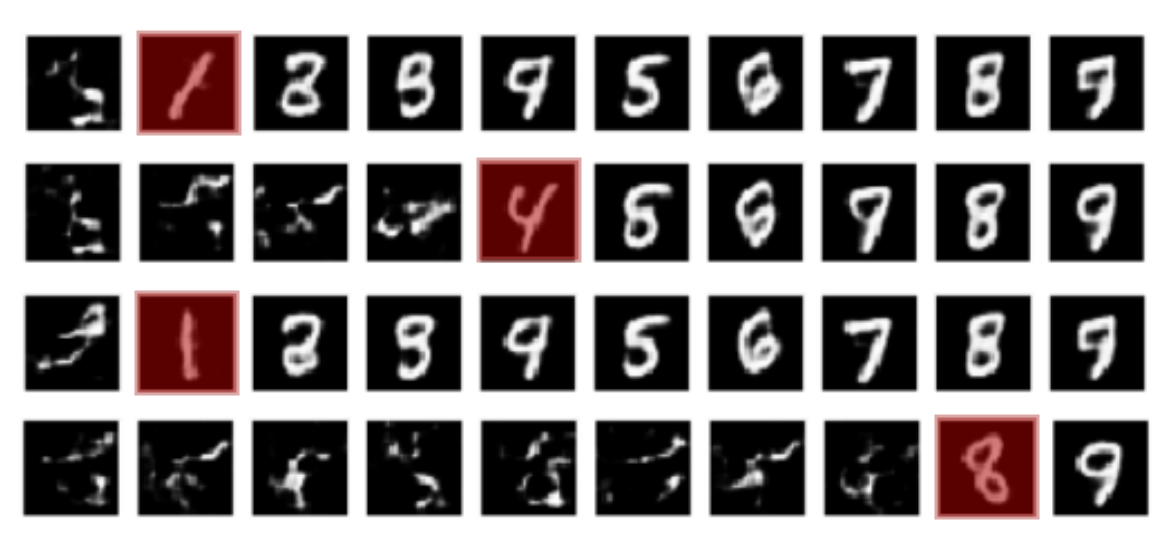}
    \caption{Without the orthogonality constraint, the inverse matrix may not exist and its approximation leads to generated images that do not resemble real digits (the application of the matrix $W$ though still leads to realistic images).
    }
    \label{fig:orth}
\end{figure}

\paragraph{Importance of orthogonality} Besides the structure preserving constraints we use another constraint in our models; the orthogonality constraint. In Fig. \ref{fig:orth} we depict the results when such constraint is not enforced in the model. While still we can generate realistic images when we apply the transformation $W$ in the latent space, the results are less ideal when we apply the inverse transformation $W^{-1}$. Since we provide no constraints on $W$ during training, it ends up being a non-singular matrix. This means that we can only approximate the inverse matrix $W^{-1}$ and, as the results indicate, this estimation is not close to the ideal. In contrast, the orthogonality constraint forces the matrix to have a well-defined inverse ($W^{-1}=W^T$).

\begin{figure*}[h]
    \centering
    \includegraphics[width=0.9\textwidth]{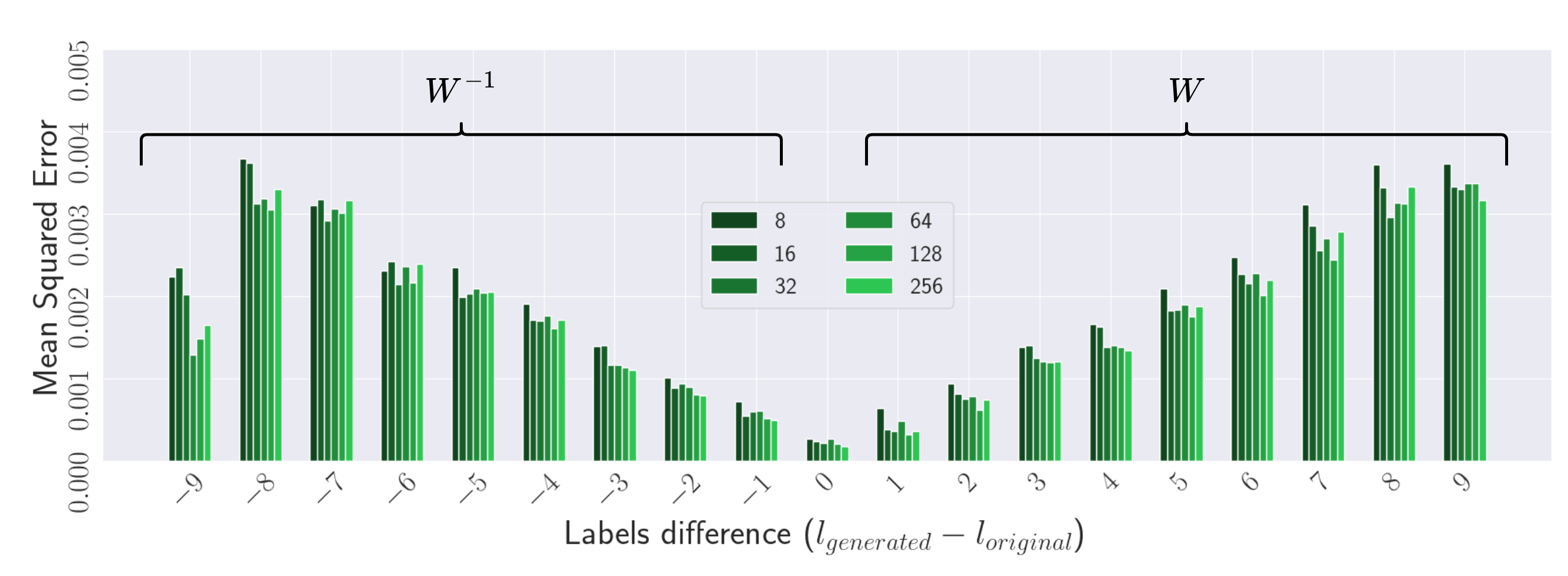}
    \caption{Mean Squared Error of the generated images for different sizes of the latent space ($n$). The application of the inverse has an only marginally higher MSE due to the orthogonality of the matrix $W$.}
    \label{fig:mse_digits}
\end{figure*}

\paragraph{Smooth image interpolation}

Since $W$ represents the abstract morphism \textit{``add one''} we can assume that the matrix $W^a$ represents the abstract morphism \textit{``add $a$''}, $\forall a\in\mathbb{R}$. This provides us with a simple way to interpolate between the digits and create smooth transitions. 

In the supplementary material we have included \textit{.gif} files that show a complete transition from $0$ to $9$ with a step of $0.2$. The results show that, although our goal was to preserve a specific relationship of the original data into the latent space, a by-product of our method is the ability of image interpolation and smooth transitions.

\subsubsection{Quantitative results}

Besides the qualitative results that indicate that our model successfully learned the relationship between the depicted digits, we can also quantify how accurate our model is. In Fig. \ref{fig:mse_digits} we show how the Mean Squared Error (MSE) changes when we continuously apply the linear transformations $W, W^{-1}$ to the latent vector and generate an image. In order to calculate it, we found the minimum MSE between the generated image and any image that depicts the specific digit. We see clearly that our model is able to perfectly generate an image of the new digit, no matter if it is smaller ($W^{-1}$ applied) or greater ($W$ applied), even after applying multiple times the linear transformations. Another important aspect is the robustness of our model to different values of $n$ (latent space dimension).

\section{Experimental evaluations - in depth}

\subsection{Dataset details}\label{sec:datasets}

The ADNI dataset can be obtained from \url{https://adni.loni.usc.edu/}. The ADNI project was launched in 2004. 

This specific dataset consists of 449 MRI scans, along with a dataset of covariates for each individual participant. Each MRI was obtained in one of three different scanners (\{GE Medical Systems, Philips Medical Systems, SIEMENS\}) and some of the covariates include \{Age, Sex, Years of Education, APOE A1, APOE A2, Marital status, Race\} along with a lot of cognitive battery scores such as \{MMSE, RAVLT, Ecog\}. 

In neuroimaging data, “APOE” typically refers to the apolipoprotein E gene. The APOE gene is involved in encoding a protein that plays a crucial role in the metabolism of lipids (fats) in the body, including cholesterol. This gene has different variants, or alleles, known as APOE $\epsilon2$, APOE $\epsilon3$, and APOE $\epsilon4$.
In the context of neuroimaging and neuroscience, the APOE gene has been of particular interest because one of its alleles, APOE $\epsilon4$, is considered a significant genetic risk factor for Alzheimer’s disease \cite{husain21, sienski21}.

The second dataset (ADCP) was shared with us by the authors of \cite{vishnu22}. 

The data for ADCP was collected through an NIH-sponsored Alzheimer’s Disease Connectome Project
(ADCP) U01 AG051216. The study inclusion criteria for AD (Alzheimer’s disease) / MCI (Mild Cognitive Impairment) patients consisted of age between 55-90 years, willing and able to undergo all procedures, retaining decisional
capacity at the initial visit, and meet criteria for probable AD or MCI. The MRI images were acquired at three distinct sites. 

Besides the medical image datasets, we evaluate our performance in two tabular datasets; the German\citep{german_credit_data} and the Adult \cite{misc_adult_2}. The German consists of $58$ features, with some of them being \textit{foreigner} and \textit{age}, and our goal is to predict the consumers' default loans. The Adult consists of $100$ with sensitive attributes such as \textit{sex} and \textit{age}. Our goal here is to predict whether someone's income is higher than \$$50$K.

\subsection{Results on tabular data}\label{sec:tabular}

Tables \ref{tab:results_german}, \ref{tab:results_adult} contains all 3 metrics for German and Adult datasets, respectively. We observe that we have accuracy similar to the other methods, while we have a significantly lower value of $\mathcal{MMD}$ and $\mathcal{ADV}$, meaning that we obtain a less biased embedding space and, as a result, predictions.

In all of our experiments we used a latent space of size $32$ and a 1-hidden-layer (of size $32$) feedforward network to map the input features to $\mathbb{R}^{32}$.

\begin{table}[t]
\centering

    \begin{tabular}{l | c: c c}
    & $\mathcal{ACC}\uparrow$ & $\mathcal{MMD}(\times 10^2)\downarrow$ & $\mathcal{ADV}\downarrow$ \\ [0.5ex] 
     \hline\hline
     \textit{Naive} & $74{\scriptstyle(0.9)}$ & $7.7{\scriptstyle(0.8)}$ & $62{\scriptstyle(3.1)}$ \\ 
     \textit{MMD}\cite{li2014learning} & $73{\scriptstyle(1.5)}$ & $1.5{\scriptstyle(0.3)}$ & $66{\scriptstyle(0.04)}$ \\
     \textit{CAI}\cite{Xie2017ControllableIT} & $76{\scriptstyle(1.3)}$ & $1.2{\scriptstyle(2.4)}$ & $65{\scriptstyle(0.01)}$ \\
     \textit{SS}\cite{pmlr-v70-zhou17c} & ${\bf 76}{\scriptstyle(0.9)}$ & $1.5{\scriptstyle(0.6)}$ & $70{\scriptstyle(6.9)}$ \\
     \textit{RM}\cite{Motiian2017UnifiedDS} & $74{\scriptstyle(2.1)}$ & $7.5{\scriptstyle(0.9)}$ & $66{\scriptstyle(4.2)}$ \\ 
     \textit{GE}\cite{vishnu22} & $75{\scriptstyle(3.3)}$ & $2.7{\scriptstyle(0.6)}$ & $54{\scriptstyle(1.1)}$ \\
     \hline

     \textbf{Ours} (inv) & $74{\scriptstyle(0.6)}$ & ${\bf 0.7}{\scriptstyle(0.1)}$ & ${\bf 51}{\scriptstyle(2.4)}$ \\
     \textbf{Ours} (1 cov) & $74{\scriptstyle(1.4)}$ & $1.3{\scriptstyle(0.9)}$ & $55{\scriptstyle(1.4)}$ \\
     \hline\hline
    \end{tabular}
    
    \caption{\textbf{Quantitative Results on German \cite{german_credit_data}.} The mean accuracy ($\mathcal{ACC}$) and invariance as evaluated by $\mathcal{MMD}$ and $\mathcal{ADV}$ are shown. The standard deviations are in parenthesis. The baseline \textit{inv} corresponds to only invariance to \textit{foreigner}, \textit{$1$ cov} corresponds to equivariance with respect to the \textit{age} covariates (plus invariance to \textit{foreigner}). $\mathcal{MMD}$ and $\mathcal{ADV}$ are significantly reduced without any significant drop in accuracy.}
    \label{tab:results_german}
\end{table}

\begin{table}
\centering

    \begin{tabular}{l | c: c c}
    & $\mathcal{ACC}\uparrow$ & $\mathcal{MMD}(\times 10^2)\downarrow$ & $\mathcal{ADV}\downarrow$ \\ [0.5ex] 
     \hline\hline
     \textit{Naive} & $84{\scriptstyle(0.1)}$ & $9.8{\scriptstyle(0.3)}$ & $83{\scriptstyle(0.1)}$ \\ 
     \textit{MMD}\cite{li2014learning} & $84{\scriptstyle(0.1)}$ & $3.1{\scriptstyle(0.3)}$ & $83{\scriptstyle(0.1)}$ \\
     \textit{CAI}\cite{Xie2017ControllableIT} & $84{\scriptstyle(0.04)}$ & $2.2{\scriptstyle(2.4)}$ & $81{\scriptstyle(0.7)}$ \\
     \textit{SS}\cite{pmlr-v70-zhou17c} &  $84{\scriptstyle(0.1)}$ & $1.5{\scriptstyle(0.2)}$ & $83{\scriptstyle(0.2)}$ \\
     \textit{RM}\cite{Motiian2017UnifiedDS} & $84{\scriptstyle(0.3)}$ & $4.8{\scriptstyle(0.7)}$ & $82{\scriptstyle(0.4)}$ \\ 
     \textit{GE}\cite{vishnu22} & $83{\scriptstyle(0.1)}$ & $7.1{\scriptstyle(0.6)}$ & $75{\scriptstyle(1.4)}$ \\
     \hline

     \textbf{Ours} (inv) & $83{\scriptstyle(0.2)}$ & ${\bf 0.6}{\scriptstyle(0.4)}$ & ${\bf 74}{\scriptstyle(1.6)}$ \\
     \textbf{Ours} (1 cov) & $84{\scriptstyle(0.1)}$ & ${\bf 1.1}{\scriptstyle(0.9)}$ & $76{\scriptstyle(1.3)}$ \\
     \hline\hline
    \end{tabular}
    
    \caption{\textbf{Quantitative Results on Adult \cite{misc_adult_2}.} The mean accuracy ($\mathcal{ACC}$) and invariance as evaluated by $\mathcal{MMD}$ and $\mathcal{ADV}$ are shown. The standard deviations are in parenthesis. The baseline \textit{inv} corresponds to only invariance to \textit{gender}, \textit{$1$ cov} corresponds to equivariance with respect to the \textit{age} covariates (plus invariance to \textit{gender}). Similarly to German, $\mathcal{MMD}$ and $\mathcal{ADV}$ are significantly reduced without any significant drop in accuracy.}
    \label{tab:results_adult}
\end{table}

\subsection{Effect of Lagrange multiplier $(\lambda)$ and latent space dimension $(n)$}\label{sec:lambda}

Here, we evaluate the effect of the latent space dimension as well as the effect of the Lagrange multiplier $\lambda$. As a reminder, the objective was defined as:

\begin{equation}
    \mathcal{L} = \underbrace{\mathcal{L}_p}_{\text{cross-entropy}} +  \sum_{s_1, s_2\in\mathcal{S}}\underbrace{\lambda\cdot \Big\|F(s_1) - F(s_2)\Big\|}_{\forall\ s_1, s_2 \text{ from different Scanners}} 
    \label{eq:inv}
\end{equation}

in the case of invariance, and as:

\begin{equation}
    \hspace{-0.1in}\mathcal{L}= \hspace{-0.1in}\underbrace{\mathcal{L}_p}_{\text{cross-entropy}} \hspace{-0.1in} +  \sum_{s_1, s_2\in\mathcal{S}}\underbrace{\lambda\Big\|W^{c_{s_1}-c_{s_2}} \cdot F(s_1)- F(s_2)\Big\|}_{c_{s_1}, c_{s_2} \text{ represent the nuisance covariate}}
    \label{eq:equiv}
\end{equation}

in the case of equivariance. 

Tables \ref{tab:adni_single_ablation}, \ref{tab:adcp_single_ablation} present the results for 1-covariate equivariance, for both datasets. We can observe that moderate values of $\lambda$ (i.e. $\simeq 0.01$) lead to a low MMD value (much lower than the existing methods in most of the cases) with no Accuracy drop. In fact, in most of the cases, the effect of invariance leads to better generalization capabilities (i.e. higher Accuracy).

Tables \ref{tab:adni_double_ablation}, \ref{tab:adcp_double_ablation} show the same results but in the case of 2-covariate equivariance. Although, in theory, such problem is harder to solve, this does not translate to lower values of MMD and Accuracy. In fact, MMD is slightly lower in this case, since we enforce invariance to each subgroup separately (i.e. a less ``aggressive'' form of invariance).

\begin{figure}
    \centering
    \includegraphics[width=0.7\linewidth]{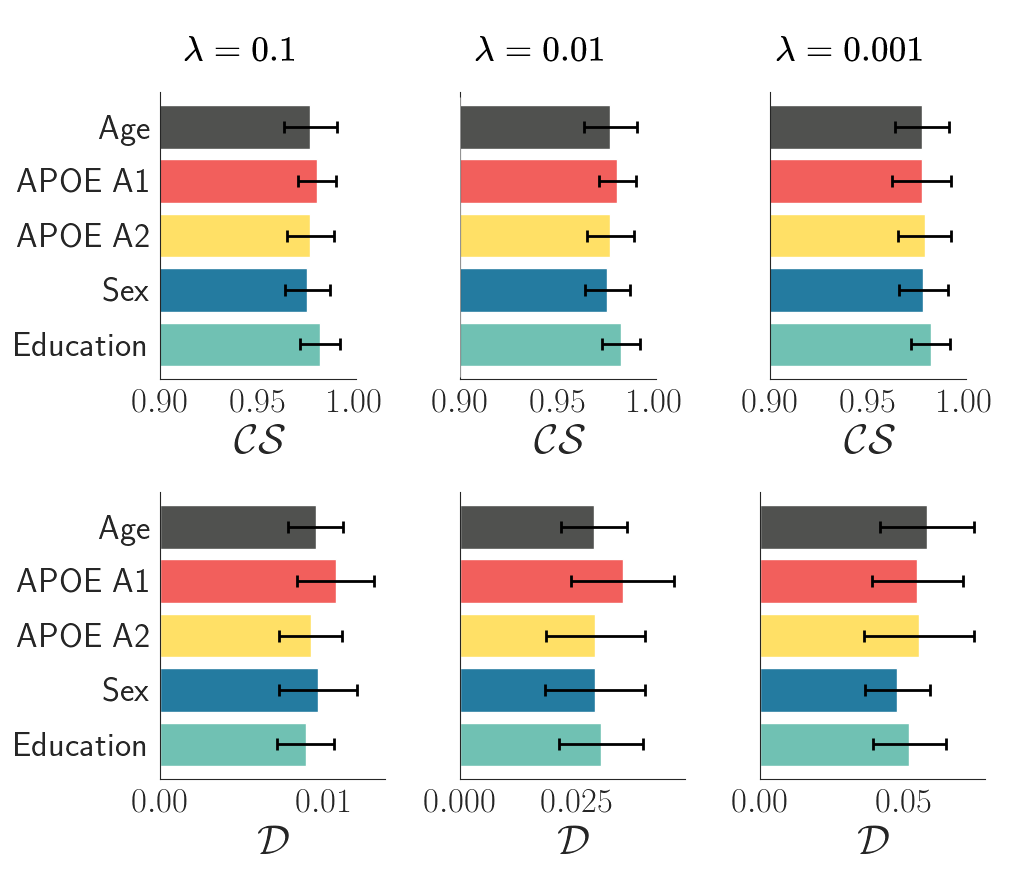}
    \caption{Effect of the Lagrange multipler $\lambda$ on the equivariance metrics $\mathcal{CS}$ and $\mathcal{D}$. As expected, $\lambda$ and $\mathcal{D}$ negatively correlated, while $\mathcal{CS}$ remains high for different values of $\lambda$.}
    \label{fig:cs_d_ablation}
\end{figure}

Finally, in Table \ref{tab:adni_5_ablation} and Fig. \ref{fig:cs_d_ablation} we present the results for 5-covariate equivariance, for the ADNI dataset. The results show, undoubtedly, that we have presented a general method which has no problem in handling multiple nuisance covariates, in contrast to existing approaches that either can not handle such scenarios, or exhibit a high performance drop.

\begin{table*}[h]
    \centering
    \begin{tabular}{l | c c c | c c c }
    
    & \multicolumn{6}{c}{Only invariance}  \\
    \hdashline
    & \multicolumn{3}{c}{$\mathcal{ACC}$} & \multicolumn{3}{c}{$\mathcal{MMD}$}  \\
    \cline{2-7}\cline{2-7}
    $\lambda$ & $16$ & $32$ & $64$ & $16$ & $32$ & $64$  \\
    \hline
    $0.001$ & $81{\scriptstyle(2.3)}$ & $81{\scriptstyle(1.3)}$ & $84{\scriptstyle(3.9)}$ & $18{\scriptstyle(4.4)}$ & $21{\scriptstyle(0.6)}$ & $24{\scriptstyle(0.8)}$  \\
    $0.01$ & $81{\scriptstyle(2.3)}$ & $79{\scriptstyle(0.1)}$ & $81{\scriptstyle(1.3)}$ & $06{\scriptstyle(2.0)}$ & $17{\scriptstyle(5.0)}$ & $16{\scriptstyle(4.8)}$  \\
    $0.1$ & $81{\scriptstyle(4.0)}$ & $78{\scriptstyle(3.8)}$ & $83{\scriptstyle(0.6)}$ & $02{\scriptstyle(2.4)}$ & $07{\scriptstyle(3.3)}$ & $13{\scriptstyle(8.4)}$  \\
    \end{tabular}
    \caption{ADNI \citep{mueller05}: Effect of the Lagrange multiplier $\lambda$ and size of the latent space $n$ to Accuracy and Maximum Mean Discrepancy (MMD), where we enforce only invariance (with respect to the site) to the latent space. Higher values of $\lambda$ lead to lower MMD (more invariant representations). In all the combinations of $(\lambda, n)$ Accuracy remains high (in most of the cases higher than the baselines) while MMD is significantly lower than the baselines in most cases.}
    \label{tab:adni_inv_ablation}
\end{table*}

\begin{table*}[h]
    \centering
    \begin{tabular}{c|l | c c c | c c c | c c c } 
    \multicolumn{2}{c}{}  & \multicolumn{3}{c|}{Age} & \multicolumn{3}{c|}{Sex} & \multicolumn{3}{c}{APOE A1}  \\
    \cline{3-11}\cline{3-11}
    \multicolumn{2}{c|}{$\lambda$} & $16$ & $32$ & $64$ & $16$ & $32$ & $64$ & $16$ & $32$ & $64$ \\
    \hline
    
    \parbox[t]{2mm}{\multirow{3}{*}{\rotatebox[origin=c]{90}{$\mathcal{ACC}$}}} & $0.001$ & $79{\scriptstyle(3.3)}$ & $80{\scriptstyle(5.7)}$ & $79{\scriptstyle(4.0)}$ & $80{\scriptstyle(3.3)}$ & $81{\scriptstyle(3.0)}$ & $80{\scriptstyle(3.4)}$ & $79{\scriptstyle(3.9)}$ & $80{\scriptstyle(1.7)}$ & $79{\scriptstyle(3.2)}$ \\
    & $0.01$ & $82{\scriptstyle(2.5)}$ & $80{\scriptstyle(3.2)}$ & $81{\scriptstyle(3.4)}$ & $80{\scriptstyle(4.5)}$ & $80{\scriptstyle(2.2)}$ & $81{\scriptstyle(1.3)}$ & $82{\scriptstyle(1.7)}$ & $80{\scriptstyle(3.4)}$ & $81{\scriptstyle(5.0)}$ \\
    & $0.1$ & $79{\scriptstyle(1.3)}$ & $79{\scriptstyle(3.2)}$ & $74{\scriptstyle(2.9)}$ & $78{\scriptstyle(3.9)}$ & $77{\scriptstyle(1.7)}$ & $77{\scriptstyle(1.7)}$ & $78{\scriptstyle(4.2)}$ & $77{\scriptstyle(1.1)}$ & $74{\scriptstyle(1.1)}$ \\
    \hline\hline
    \parbox[t]{2mm}{\multirow{3}{*}{\rotatebox[origin=c]{90}{$\mathcal{MMD}$}}} & $0.001$ & $27{\scriptstyle(4.4)}$ & $27{\scriptstyle(0.6)}$ & $27{\scriptstyle(0.8)}$ & $27{\scriptstyle(1.2)}$ & $26{\scriptstyle(1.0)}$ & $27{\scriptstyle(3.6)}$ & $28{\scriptstyle(0.7)}$ & $27{\scriptstyle(0.7)}$ & $27{\scriptstyle(0.1)}$ \\
    & $0.01$ & $11{\scriptstyle(2.9)}$ & $15{\scriptstyle(6.5)}$ & $19{\scriptstyle(3.6)}$ & $17{\scriptstyle(6.3)}$ & $13{\scriptstyle(2.9)}$ & $20{\scriptstyle(6.1)}$ & $11{\scriptstyle(4.7)}$ & $15{\scriptstyle(2.8)}$ & $22{\scriptstyle(4.0)}$ \\
    & $0.1$ & $04{\scriptstyle(0.7)}$ & $09{\scriptstyle(8.1)}$ & $16{\scriptstyle(2.5)}$ & $03{\scriptstyle(0.4)}$ & $07{\scriptstyle(6.0)}$ & $15{\scriptstyle(5.1)}$ & $04{\scriptstyle(0.6)}$ & $11{\scriptstyle(4.4)}$ & $14{\scriptstyle(1.4)}$ \\
    \end{tabular}
    \caption{ADNI \citep{mueller05}: Effect of the Lagrange multiplier $\lambda$ and size of the latent space $n$ to Accuracy and to Maximum Mean Discrepancy (MMD) in the case of a single covariate equivariance. No Accuracy drop is observed, while MMD is significantly lower than the baselines, in most of the experiments.}
    \label{tab:adni_single_ablation}
\end{table*}

\begin{table*}[h]
    \centering
    \begin{tabular}{c|l | c c c | c c c | c c c } 
    \multicolumn{2}{c}{}  & \multicolumn{3}{c|}{Age-APOE A1} & \multicolumn{3}{c|}{Age-Sex} & \multicolumn{3}{c}{APOE A1-Sex}  \\
    \cline{3-11}\cline{3-11}
    \multicolumn{2}{c|}{$\lambda$} & $16$ & $32$ & $64$ & $16$ & $32$ & $64$ & $16$ & $32$ & $64$ \\
    \hline
    \parbox[t]{2mm}{\multirow{3}{*}{\rotatebox[origin=c]{90}{$\mathcal{ACC}$}}} & $0.001$ & $80{\scriptstyle(1.1)}$ & $79{\scriptstyle(3.2)}$ & $80{\scriptstyle(3.4)}$ & $82{\scriptstyle(4.2)}$ & $82{\scriptstyle(2.3)}$ & $82{\scriptstyle(2.2)}$ & $83{\scriptstyle(1.3)}$ & $81{\scriptstyle(2.2)}$ & $80{\scriptstyle(2.8)}$ \\
    & $0.01$ & $80{\scriptstyle(1.3)}$ & $81{\scriptstyle(1.7)}$ & $82{\scriptstyle(1.7)}$ & $80{\scriptstyle(1.1)}$ & $81{\scriptstyle(3.6)}$ & $82{\scriptstyle(3.0)}$ & $82{\scriptstyle(2.2)}$ & $79{\scriptstyle(3.0)}$ & $80{\scriptstyle(3.6)}$ \\
    & $0.1$ & $78{\scriptstyle(2.2)}$ & $77{\scriptstyle(1.9)}$ & $75{\scriptstyle(2.8)}$ & $77{\scriptstyle(2.8)}$ & $77{\scriptstyle(0.6)}$ & $77{\scriptstyle(2.3)}$ & $77{\scriptstyle(4.2)}$ & $79{\scriptstyle(2.3)}$ & $76{\scriptstyle(0.6)}$ \\
    \hline\hline
    \parbox[t]{2mm}{\multirow{3}{*}{\rotatebox[origin=c]{90}{$\mathcal{MMD}$}}} & $0.001$ & $24{\scriptstyle(2.8)}$ & $26{\scriptstyle(0.4)}$ & $27{\scriptstyle(0.1)}$ & $22{\scriptstyle(1.1)}$ & $27{\scriptstyle(2.9)}$ & $27{\scriptstyle(0.4)}$ & $25{\scriptstyle(3.7)}$ & $26{\scriptstyle(8.0)}$ & $26{\scriptstyle(4.0)}$ \\
    & $0.01$ & $07{\scriptstyle(1.2)}$ & $14{\scriptstyle(7.0)}$ & $18{\scriptstyle(5.4)}$ & $09{\scriptstyle(0.4)}$ & $17{\scriptstyle(3.9)}$ & $16{\scriptstyle(4.7)}$ & $10{\scriptstyle(6.6)}$ & $16{\scriptstyle(5.7)}$ & $15{\scriptstyle(6.0)}$ \\
    & $0.1$ & $04{\scriptstyle(4.7)}$ & $07{\scriptstyle(0.9)}$ & $10{\scriptstyle(7.0)}$ & $2{\scriptstyle(0.7)}$ & $04{\scriptstyle(2.2)}$ & $08{\scriptstyle(3.4)}$ & $03{\scriptstyle(2.6)}$ & $09{\scriptstyle(9.0)}$ & $09{\scriptstyle(4.0)}$ \\
    \end{tabular}
    \caption{ADNI \citep{mueller05}: Effect of the Lagrange multiplier $\lambda$ and size of the latent space $n$ to Accuracy and to Maximum Mean Discrepancy (MMD) in the case of a two-covariate equivariance.}
    \label{tab:adni_double_ablation}
\end{table*}

\begin{table*}[h]
    \centering
    \begin{tabular}{l | c c c | c c c }
    
    & \multicolumn{6}{c}{Age-Sex-APOE A1-APOE A2-Education}  \\
    \hdashline
    & \multicolumn{3}{c}{$\mathcal{ACC}$} & \multicolumn{3}{c}{$\mathcal{MMD}$}  \\
    \cline{2-7}\cline{2-7}
    $\lambda$ & $16$ & $32$ & $64$ & $16$ & $32$ & $64$  \\
    \hline
    $0.001$ & $82{\scriptstyle(4.2)}$ & $81{\scriptstyle(1.7)}$ & $82{\scriptstyle(1.1)}$ & $11{\scriptstyle(5.3)}$ & $21{\scriptstyle(2.5)}$ & $25{\scriptstyle(3.4)}$  \\
    $0.01$ & $80{\scriptstyle(2.3)}$ & $78{\scriptstyle(2.3)}$ & $77{\scriptstyle(1.7)}$ & $05{\scriptstyle(5.3)}$ & $09{\scriptstyle(8.4)}$ & $14{\scriptstyle(3.7)}$  \\
    $0.1$ & $78{\scriptstyle(1.9)}$ & $76{\scriptstyle(2.2)}$ & $78{\scriptstyle(0.6)}$ & $01{\scriptstyle(0.2)}$ & $03{\scriptstyle(0.4)}$ & $07{\scriptstyle(5.6)}$  \\
    \end{tabular}
    \caption{ADNI \citep{mueller05}: Effect of the Lagrange multiplier $\lambda$ and size of the latent space $n$ to Accuracy and Maximum Mean Discrepancy (MMD), in the case of equivariance with respect to five variables.}
    \label{tab:adni_5_ablation}
\end{table*}

\begin{table*}[h]
    \centering
    \begin{tabular}{l | c c c | c c c }
    
    & \multicolumn{6}{c}{Only invariance}  \\
    \hdashline
    & \multicolumn{3}{c}{$\mathcal{ACC}$} & \multicolumn{3}{c}{$\mathcal{MMD}$}  \\
    \cline{2-7}\cline{2-7}
    $\lambda$ & $16$ & $32$ & $64$ & $16$ & $32$ & $64$  \\
    \hline
    $0.001$ & $82{\scriptstyle(2.2)}$ & $82{\scriptstyle(2.2)}$ & $83{\scriptstyle(4.4)}$ & $48{\scriptstyle(5.9)}$ & $53{\scriptstyle(4.8)}$ & $60{\scriptstyle(7.0)}$  \\
    $0.01$ & $83{\scriptstyle(2.2)}$ & $86{\scriptstyle(8.0)}$ & $885{\scriptstyle(3.9)}$ & $35{\scriptstyle(5.1)}$ & $34{\scriptstyle(1.2)}$ & $54{\scriptstyle(4.1)}$  \\
    $0.1$ & $81{\scriptstyle(3.8)}$ & $82{\scriptstyle(5.9)}$ & $85{\scriptstyle(7.7)}$ & $15{\scriptstyle(2.5)}$ & $25{\scriptstyle(2.2)}$ & $28{\scriptstyle(2.0)}$  \\
    \end{tabular}
    \caption{ADCP: Effect of the Lagrange multiplier $\lambda$ and size of the latent space $n$ to Accuracy and Maximum Mean Discrepancy (MMD), where we enforce only invariance (with respect to the site) to the latent space.}
    \label{tab:adcp_inv_ablation}
\end{table*}

\begin{table*}[h]
    \centering
    \begin{tabular}{c|l | c c c | c c c } 
    \multicolumn{2}{c}{}  & \multicolumn{3}{c|}{Age} & \multicolumn{3}{c}{Sex}  \\
    \cline{3-8}\cline{3-8}
    \multicolumn{2}{c|}{$\lambda$} & $16$ & $32$ & $64$ & $16$ & $32$ & $64$  \\
    \hline
    
    \parbox[t]{2mm}{\multirow{3}{*}{\rotatebox[origin=c]{90}{$\mathcal{ACC}$}}} & $0.001$ & $85{\scriptstyle(3.8)}$ & $83{\scriptstyle(2.2)}$ & $86{\scriptstyle(5.9)}$ & $83{\scriptstyle(4.4)}$ & $83{\scriptstyle(4.4)}$ & $83{\scriptstyle(5.8)}$  \\
    & $0.01$ & $79{\scriptstyle(2.2)}$ & $82{\scriptstyle(8.0)}$ & $86{\scriptstyle(5.9)}$ & $82{\scriptstyle(4.4)}$ & $81{\scriptstyle(6.7)}$ & $85{\scriptstyle(3.8)}$ \\
    & $0.1$ & $77{\scriptstyle(3.9)}$ & $79{\scriptstyle(4.4)}$ & $83{\scriptstyle(4.4)}$ & $81{\scriptstyle(0.2)}$ & $78{\scriptstyle(8.0)}$ & $83{\scriptstyle(4.4)}$ \\
    \hline\hline
    \parbox[t]{2mm}{\multirow{3}{*}{\rotatebox[origin=c]{90}{$\mathcal{MMD}$}}} & $0.001$ & $37{\scriptstyle(4.0)}$ & $51{\scriptstyle(5.6)}$ & $57{\scriptstyle(5.3)}$ & $37{\scriptstyle(4.8)}$ & $52{\scriptstyle(8.4)}$ & $55{\scriptstyle(3.0)}$ \\
    & $0.01$ & $30{\scriptstyle(4.2)}$ & $35{\scriptstyle(3.3)}$ & $39{\scriptstyle(3.6)}$ & $30{\scriptstyle(5.0)}$ & $31{\scriptstyle(4.2)}$ & $34{\scriptstyle(5.8)}$ \\
    & $0.1$ & $27{\scriptstyle(3.2)}$ & $27{\scriptstyle(5.8)}$ & $23{\scriptstyle(5.4)}$ & $23{\scriptstyle(5.3)}$ & $21{\scriptstyle(7.0)}$ & $46{\scriptstyle(4.2)}$ \\
    \end{tabular}
    \caption{ADCP: Effect of the Lagrange multiplier $\lambda$ and size of the latent space $n$ to Accuracy and to Maximum Mean Discrepancy (MMD) in the case of a single covariate equivariance.}
    \label{tab:adcp_single_ablation}
\end{table*}

\begin{table*}
    \centering
    \begin{tabular}{l | c c c | c c c }
    
    & \multicolumn{6}{c}{Age-Sex}  \\
    \hdashline
    & \multicolumn{3}{c}{$\mathcal{ACC}$} & \multicolumn{3}{c}{$\mathcal{MMD}$}  \\
    \cline{2-7}\cline{2-7}
    $\lambda$ & $16$ & $32$ & $64$ & $16$ & $32$ & $64$  \\
    \hline
    $0.001$ & $83{\scriptstyle(5.8)}$ & $82{\scriptstyle(4.4)}$ & $82{\scriptstyle(9.6)}$ & $37{\scriptstyle(2.1)}$ & $42{\scriptstyle(9.4)}$ & $49{\scriptstyle(4.1)}$  \\
    $0.01$ & $85{\scriptstyle(7.7)}$ & $82{\scriptstyle(9.5)}$ & $85{\scriptstyle(3.8)}$ & $29{\scriptstyle(3.3)}$ & $32{\scriptstyle(1.2)}$ & $35{\scriptstyle(6.1)}$  \\
    $0.1$ & $79{\scriptstyle(2.2)}$ & $79{\scriptstyle(8.7)}$ & $83{\scriptstyle(4.4)}$ & $24{\scriptstyle(4.6)}$ & $27{\scriptstyle(4.8)}$ & $22{\scriptstyle(2.8)}$  \\
    \end{tabular}
    \caption{ADCP: Effect of the Lagrange multiplier $\lambda$ and size of the latent space $n$ to Accuracy and Maximum Mean Discrepancy (MMD), in the case of two-covariates equivariance.}
    \label{tab:adcp_double_ablation}
\end{table*}

\end{document}